\documentclass{article}

\usepackage[final]{corl_2025} % Uncomment for the camera-ready ``final'' version.
\usepackage{graphicx}
\usepackage{multicol}
\usepackage{booktabs}
\usepackage{amsmath}
\usepackage{multirow}
\usepackage{amssymb}
\usepackage{subcaption}
\usepackage[capitalize]{cleveref}
\usepackage{wrapfig}
\usepackage{enumitem}

\newcommand{\ob}{\mathbf{o}}
\newcommand{\ac}{\mathbf{a}}
\newcommand{\z}{\mathbf{z}}

\newcommand{\custom}{tabletop}
\newcommand{\pair}{scene-aligned cross-embodiment dataset}

\newcommand\blfootnote[1]{%
  \begingroup
  \renewcommand\thefootnote{}\footnote{#1}%
  \addtocounter{footnote}{-1}%
  \endgroup
}

\title{UniSkill: Imitating Human Videos via \\Cross-Embodiment Skill Representations}

\author{
  Hanjung Kim$^\ast$ \hspace{4mm} Jaehyun Kang$^\ast$ \hspace{4mm} Hyolim Kang \hspace{4mm} Meedeum Cho\\
  \textbf{Seon Joo Kim} \hspace{4mm} \textbf{Youngwoon Lee}\\[0.3cm]
  Yonsei University\\[0.3cm]
  \url{https://kimhanjung.github.io/UniSkill}
}

\begin{document}
\maketitle

%===============================================================================

\begin{abstract}
    Mimicry is a fundamental learning mechanism in humans, enabling individuals to learn new tasks by observing and imitating experts. However, applying this ability to robots presents significant challenges due to the inherent differences between human and robot embodiments in both their visual appearance and physical capabilities. While previous methods bridge this gap using cross-embodiment datasets with shared scenes and tasks, collecting such aligned data between humans and robots at scale is not trivial. In this paper, we propose UniSkill, a novel framework that learns embodiment-agnostic skill representations from large-scale cross-embodiment video data without any labels, enabling skills extracted from human video prompts to effectively transfer to robot policies trained only on robot data. Our experiments in both simulation and real-world environments show that our cross-embodiment skills successfully guide robots in selecting appropriate actions, even with unseen video prompts. The project website can be found at: \url{https://kimhanjung.github.io/UniSkill}.
\end{abstract}

% Two or three meaningful keywords should be added here

\blfootnote{$^\ast$ denotes equal contributions.}

\section{Introduction}
\label{sec:introduction}

Learning from human videos has emerged as a central paradigm in robot learning, offering a scalable approach to the scarcity of robot-specific data by leveraging large, diverse video sources.
Human videos contain everyday behaviors such as human-object interactions, which could provide a rich source of skills for robot learning. Here, a central question arises: \textit{Can robots acquire cross-embodiment skill representations by watching large-scale human demonstrations?}

Translating human videos into robot-executable skill representations has traditionally relied on paired human-robot datasets~\citep{yu2018one, bc-z, vid2robot} or predefined semantic skill labels~\citep{star, vip-prompt}, both of which are difficult to scale. 
Recent approaches aim to bypass these requirements by learning cross-embodiment skill representations without explicit pairing or labeling~\citep{xskill, mimicplay, kareer2024egomimic, wen2024any, im2flow2act}. 
However, these methods still impose constraints on data collection, such as multi-view camera setups, and task and scene alignment between human and robot demonstrations, which limit their scalability and applicability to real-world, in-the-wild human videos.

\begin{figure}[t]
    \centering
    \includegraphics[width=\linewidth]{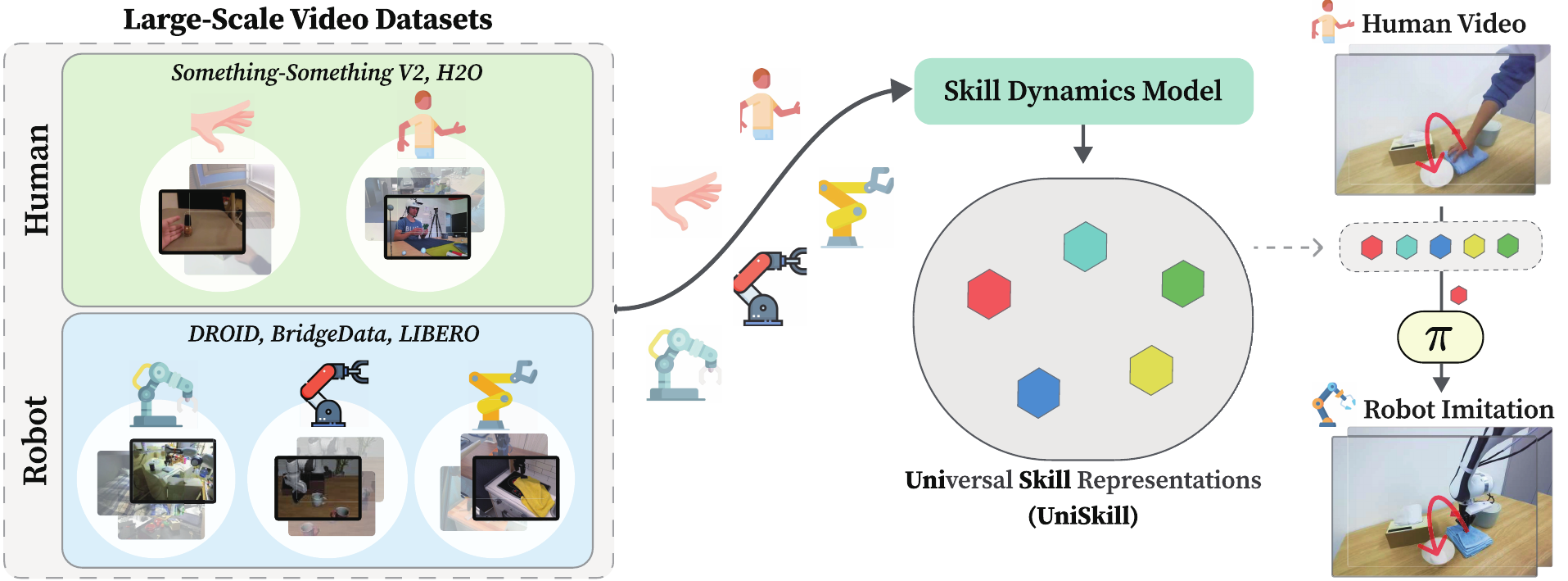}
    \caption{\textbf{Uni}versal \textbf{Skill} representations (\textbf{UniSkill}) are cross-embodiment skill representations shared across various embodiments (e.g., humans, Franka, WidowX) trained from both human and robot videos via skill dynamics modeling. Unlike prior works that require additional supervision (e.g., trajectory labels) or alignment between human and robot videos, UniSkill removes these constraints by learning solely from off-the-shelf video datasets--such as Something-Something V2~\citep{sthsth} and H2O~\citep{h2o} for human videos, and DROID~\citep{droid}, Bridge V2~\citep{bridge}, and LIBERO~\citep{libero} for robot videos. UniSkill, trained on large-scale cross-embodiment videos, learns an embodiment-agnostic skill representation that enables interpreting human videos as skill sequences executable directly through a skill-conditioned policy.
    }
    \label{fig:teaser}
\end{figure}

To this end, we propose \textbf{Uni}versal \textbf{Skill} representations (\textbf{UniSkill}), a scalable approach for learning cross-embodiment skill representations \textit{from large-scale in-the-wild video data} so that a robot can translate \textit{an unseen human demonstration} into a sequence of robot-executable skill representations, as illustrated in \Cref{fig:teaser}. 
To extract reusable, embodiment-agnostic motion patterns from videos, UniSkill focuses on capturing dynamics changes between temporally distant video frames, which can be agnostic to embodiments and shared across diverse videos.
UniSkill leverages an image-editing pipeline, which naturally emphasizes dynamic regions over static content, and encodes the resulting motion patterns into skill representations.
The design choice enables the use of arbitrary, embodiment-agnostic video datasets for training, making it possible to scale cross-embodiment skill representation learning to large, in-the-wild datasets.
As a result of its embodiment-agnostic skill representation, UniSkill can imitate a given prompt video by capturing sequence of motion patterns within it, even when demonstration is performed by a human.

% Teaser

Our experiments demonstrate that UniSkill effectively learn cross-embodiment skill representations by training on large-scale video datasets.
Its embodiment-agnostic design allows it to generalize to unseen human prompts at test time, without any kind of additional guidance such as language instructions.
Notably, UniSkill's skill-centric architecture enhances robustness to novel objects and supports compositional task solving.
In addition, its versatile training pipeline benefits from incorporating diverse video datasets, with performance improving as more data sources are added.
Finally, qualitative results from the Forward Skill Dynamics (FSD) model predictions and skill representation visualizations highlight the interpretability of the learned representations.

In summary, our contributions are twofold:
\begin{itemize}[leftmargin=2em]
    \item We introduce UniSkill, a universal skill representation learning approach that enables the use of large-scale video data by removing the need for labels or any form of alignment constraints.
    \item UniSkill shows effective human-to-robot and robot-to-robot imitation in both simulation and real-world experiments through its embodiment-agnostic skill representation.
\end{itemize}

\section{Related Work}
\label{sec:related_work}

Learning action (or skill) representations for robot learning from in-the-wild video dataset is challenging due to the absence of action labels. Recent work on \textbf{latent action models} addresses this by deriving action-relevant information through inverse or forward dynamics models. LAPO~\citep{lapo} and Genie~\citep{genie} propose to learn generative interactive environments from gameplay videos with latent actions, but they are primarily tailored to game settings with discrete actions. LAPA~\citep{lapa} extends this line of research to real-world robotic manipulation by incorporating diverse videos, including human demonstrations. However, the learned latent actions are used merely to pretrain policy as pseudo action labels. Going one step further, UniSkill treats latent actions as explicit skill representations and directly trains a skill-conditioned policy on the learned representations.

Another line of work transfers action information from human videos to robots via explicit action representations, such as 2D/3D trajectories and flow fields. MimicPlay~\citep{mimicplay}, EgoMimic~\citep{kareer2024egomimic}, and Motion Tracks~\citep{ren2025motion} extract 3D human hand trajectories from multi-view videos or wearable sensor inputs. ATM~\citep{wen2024any} and Im2Flow2Act~\citep{im2flow2act} predict 2D motion paths or flows from task-labeled human videos. These methods often require calibrated cameras, pose tracking, or environment-specific constraints, limiting their scalability to off-the-shelf video datasets.
\textit{UniSkill differs by avoiding any task-specific trajectory extraction or pose supervision.} Our method learns directly from raw RGB videos, which enables the use of diverse public human and robot datasets.

XSkill~\citep{xskill} is the most similar work to our paper, as XSkill does not rely on manually designed skill representations or supervision, such as hand trajectories or flows. XSkill aligns skills from human and robot videos via Sinkhorn-Knopp clustering~\citep{cuturi2013sinkhorn, caron2020unsupervised}, enforcing embodiment-agnostic skill prototypes. However, this clustering with shared prototypes implicitly assumes some degree of alignment between human and robot videos. In practice, while paired dataset is not required, human videos still cover the target robot task and be captured in similar environments for effective skill transfer. 
On the other hand, UniSkill takes a different approach, \textit{learning predictive representations through future frame forecasting.} This completely removes the need for domain or task alignment, allowing the model to benefit even from entirely unrelated human videos. As a result, UniSkill can fully exploit web-scale, unlabeled data for cross-embodiment skill representation learning.

\section{Method}
\label{sec:method}

\begin{figure}[t]
    \centering
    \includegraphics[width=\textwidth]{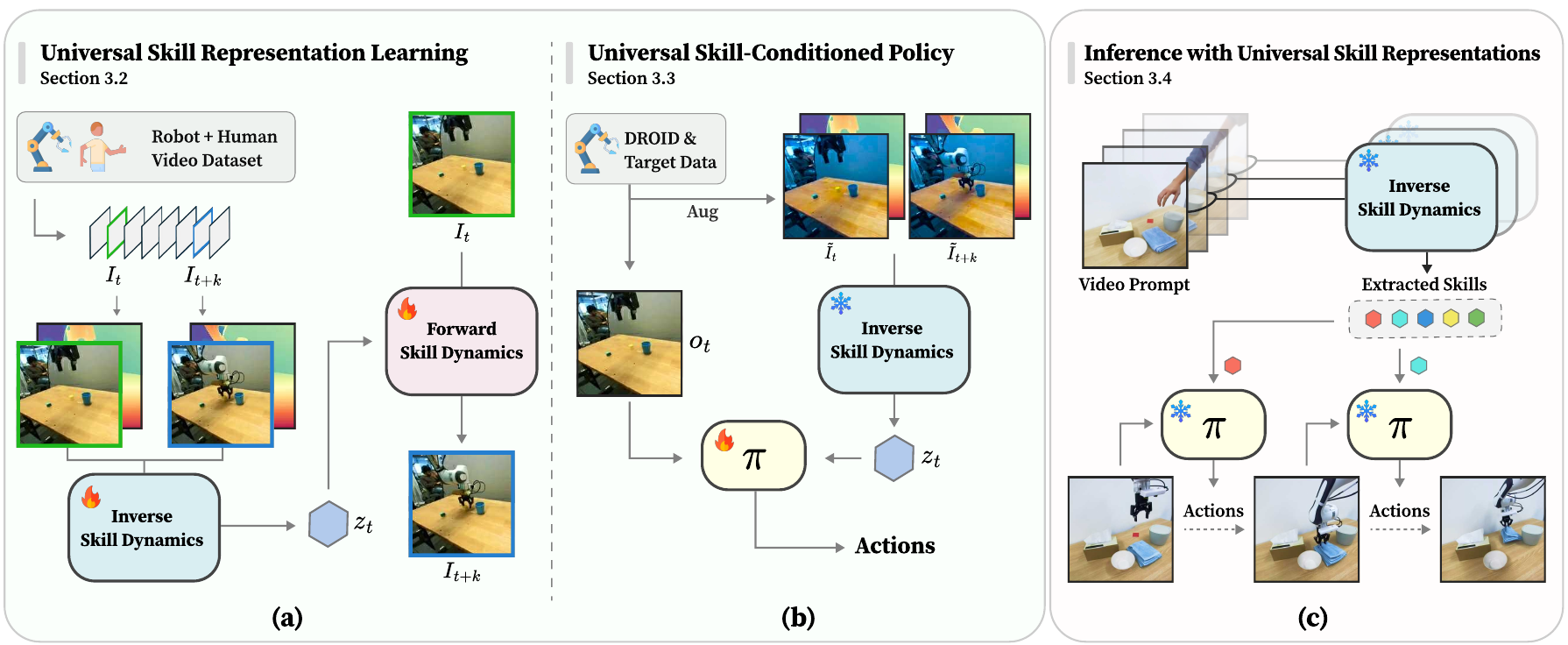}
    \caption{The overview of UniSkill. (a)~Inverse Skill Dynamics (ISD) and Forward Skill Dynamics (FSD) are jointly trained on diverse video datasets to encode dynamics information into universal skill representations by predicting skills and future frames, respectively. (b)~A universal skill-conditioned policy is trained on DROID and small target environment data. Here, skill representations are extracted from robot data using the pretrained ISD. (c)~Skills extracted from a human video prompt are sequentially executed by the skill-conditioned policy to reproduce the target behavior.}
    \label{fig:mainfigure}
\end{figure}

In this paper, we address the problem of cross-embodiment imitation, where a human guides a robot to perform a task by demonstrating the desired behavior.
We introduce UniSkill, which learns embodiment-agnostic skill representations from large-scale, unlabeled video data spanning diverse embodiments (\Cref{sec:method:skill_representation}), and imitates a human video demonstration through a skill-conditioned robot policy (\Cref{sec:method:policy}) and cross-embodiment skills extracted from the video demonstration (\Cref{sec:method:inference}), as illustrated in \Cref{fig:mainfigure}.

% Main Figure

\subsection{Problem Formulation}
\label{sec:method:problem}

We aim for cross-embodiment imitation, where a skill-conditioned robot policy $\pi(\ob_t, \z_t)$ replicates behaviors demonstrated in a prompt video $\mathcal{V}^p = \{I_1^p, \dots, I_{N_p}^p\}$ of length $N_p$, which comes from a different embodiment (e.g., a human).
$I_t^p$ and $\ob_t$ represent the frame of the prompt video and the robot observation at time $t$, respectively. The prompt video contains only raw pixel data, without any action annotations.
To achieve imitation, we extract an embodiment-agnostic skill representation $\z_t$ from a pair of frames $(I_t^p, I_{t+k}^p)$ within the prompt video, where $k$ is the temporal distance between frames. This skill representation $\z_t$ is then used to condition the robot’s policy $\pi(\ob_t, \z_t)$, enabling it to replicate the actions demonstrated in the video prompt.

For training, we assume two types of datasets: (1)~cross-embodiment video datasets $\mathcal{D}_u = \{\mathcal{V}^u_n\}_{n=1}^{N_u}$ and (2)~robot demonstration datasets $\mathcal{D}_a = \{\mathcal{T}_n\}_{n=1}^{N_a}$.
First, an unlabeled large-scale video dataset $\mathcal{D}_u$ consists of both human and robot videos, where each video $\mathcal{V}^u$ contains only raw RGB frames $I^u$. Then, a robot dataset consists of action-labeled trajectories, where each trajectory $\mathcal{T}_n$ is a sequence of observation-action pairs: $\mathcal{T}_n = \{(\ob_t, \ac_t)\}_{t=1}^{L_n}$.
$L_n$ denotes the length of the $n$-th trajectory, and $\ob_t$ and $\ac_t$ represent the observation and the corresponding robot action at time $t$, respectively.
$\mathcal{D}_a$ is relatively smaller than $\mathcal{D}_u$ (i.e., $N_u \gg N_a$).
Unless otherwise stated, $\mathcal{V}_p$ is excluded from both $\mathcal{D}_u$ and $\mathcal{D}_a$, ensuring that the prompt videos remain unseen during training.

\subsection{Universal Skill Representation Learning}
\label{sec:method:skill_representation}
Following~\cite{lapo}, we develop UniSkill based on the intuition that the latent skill $\z_t$ serves as an effective compression of the dynamics between $I_t$ and $I_{t+k}$, thus using the reconstruction of $I_{t+k}$ as a supervisory signal.  
In addition, we impose an additional requirement: the extracted $\z_t$ should be embodiment-agnostic. 
In other words, if the semantic meanings of the dynamics are the same, the extracted skills should be similar, regardless of the actor.  
Therefore, we fully leverage the embodiment-agnostic nature of motion patterns in videos by introducing an \textit{Inverse Skill Dynamics} (ISD) model and a \textit{Forward Skill Dynamics} (FSD) model, trained on a large-scale, multi-embodiment, unlabeled video dataset $D_u$.

\textbf{Inverse Skill Dynamics Model (ISD)} consumes two temporally distant frames $I_t$ and $I_{t+k}$, and yields a universal skill representation $\z_t$, namely:
\begin{equation}
\label{eq:enc}
    \z_t=ISD(I_t, I_{t+k}).
\end{equation}
We found that relying solely on raw RGB frames can lead to encoding of embodiment-specific details, such as the demonstrator’s appearance or scene context, which can hinder the learning of embodiment-agnostic $\z_t$.  
To mitigate this, as illustrated in \Cref{fig:mainfigure} (a), we incorporate depth information by generating depth maps for each frame using an off-the-shelf monocular depth estimator~\cite{yang2024depth}.  
Note that we do not use external depth inputs; instead, our ISD model internally employs a depth estimation module, utilizing predicted depth as an intermediate representation.
Further analysis of depth utilization is provided in \Cref{sec:experiments:ablation}.

\textbf{Forward Skill Dynamics Model (FSD)} predicts the future frame $I_{t+k}$ given $I_t$ and $\z_t$:
\begin{equation}
\label{eq:dec}
    I_{t+k} = FSD(I_t, \z_t).
\end{equation}
To prevent a trivial solution where FSD simply assigns $\z_t = I_{t+k}$, we enforce an information bottleneck on $\z_t$, following ~\cite{lapo}.  
Since $I_t$ and $I_{t+k}$ belong to the same video and are only $k$ frames apart, the dynamics may induce minimal changes to the overall scene, except for the embodiment and its relevant parts.  
Thus, we formulate the prediction process as an image editing task, modifying only the dynamic components while preserving the rest of the scene.  
To be specific, we adopt a diffusion-based image editing method, InstructPix2Pix~\cite{instructpix2pix}, which generates a target image from a source image using a language instruction.  
In our framework, we replace the language instruction with $\z_t$, enabling the future frame to be generated according to the skill representation.  
Following InstructPix2Pix~\cite{instructpix2pix}, we minimize the latent diffusion objective during training, encouraging ISD to compactly encode the dynamic information to $\z_t$.

\subsection{Universal Skill-Conditioned Policy}
\label{sec:method:policy}

The next stage involves training a robot policy network $\pi_\phi(\ac_{t:t+h} \mid \ob_t, \z_t)$, which receives the current observation $\ob_t$ and utilizes $\z_t$ as a skill-conditioning signal.
To train the skill-conditioned policy, we first sample two observations, $\ob_t$ and $\ob_{t+k}$, from $\mathcal{D}_a$ and extract the skill representation $\z_t=ISD(I_t, I_{t+k})$ using the pre-trained, frozen ISD.  
The policy $\pi_{\phi}$ is then conditioned on $\ob_t$ and $\z_t$ to predict a sequence of actions $\ac_{t:t+h}$, where $h$ denotes the action horizon~\citep{diffusionpolicy, aloha}. Finally, the policy is trained using behavioral cloning on a robot dataset $\mathcal{D}_a$:
\begin{equation} \label{eq:policy}
     \phi^*=\text{argmax}_{\phi} \mathbb{E}_{(\ob_t, \ob_{t+h}, \ac_{t:t+h}) \sim \mathcal{D}_a} \left[ \log \pi_{\phi}(\ac_{t:t+h} \mid \ob_t, \z_t) \right].
\end{equation}

For cross-embodiment imitation, the policy receives $\z_t$ from videos with different embodiments at inference time while the policy is trained solely on $\z_t$ computed from robot videos.
To mitigate the discrepancy between $\z_t$ for training and testing, we apply augmentation to both $I_t$ and $I_{t+k}$, producing $\tilde{I}_t$ and $\tilde{I}_{t+k}$ to simulate the aforementioned discrepancy during training, as illustrated in \Cref{fig:mainfigure} (b).  
This augmentation enhances the robustness of our skill-conditioned policy, enabling it to generalize effectively across diverse video prompts $\mathcal{V}_p$ from different embodiment at inference time.  
The effectiveness of this augmentation is demonstrated in \Cref{sec:experiments:ablation}.

\subsection{Cross-Embodiment Imitation with Universal Skill Representations}
\label{sec:method:inference}

During inference, the behaviors demonstrated in the video prompts are imitated using the frozen ISD and skill-conditioned policy $\pi_\phi$.
Given a video prompt $\mathcal{V}_p$, we extract a set of skill representations $\{\z_i\}_{i=1}^{N_z}$, where $N_z< N_p$, using ISD.
As shown in \Cref{fig:mainfigure} (c), we sequentially condition the policy $\pi_\phi$ on each skill representation $\z_i$ to predict the corresponding actions that imitate the demonstrated behaviors in $\mathcal{V}_p$.
Importantly, the universal skill representation learned by ISD allows $\pi_\phi$ to condition on video prompts from any embodiment.

\section{Experiments}
\label{sec:experiments}

\subsection{Experimental Setup}

\paragraph{Datasets.}
Our primary goal is to learn the underlying dynamics from large video datasets across diverse embodiments.
Thus, we leverage a variety of video domains, including human and robot videos from both real-world and simulated environments:
\begin{itemize}[leftmargin=2em]
    \item \textbf{Human video datasets}: Something-Something V2~\cite{sthsth} and H2O~\cite{h2o}
    \item \textbf{Robot video datasets}: DROID~\cite{droid}, BridgeV2~\cite{bridge}, and LIBERO~\cite{libero}
\end{itemize}
Something-Something V2 contains numerous clips of humans performing simple actions in ego-centric view and H2O includes both ego-centric and third-person viewpoints, featuring two-handed manipulation.
DROID and BridgeV2 are large-scale manipulation datasets, featuring a Franka robot arm and a WidowX 250 arm, respectively. LIBERO is a simulation dataset in which a Franka arm performs various tasks in diverse environments.

\begin{figure}[t]
    \centering
    \includegraphics[width=\textwidth]{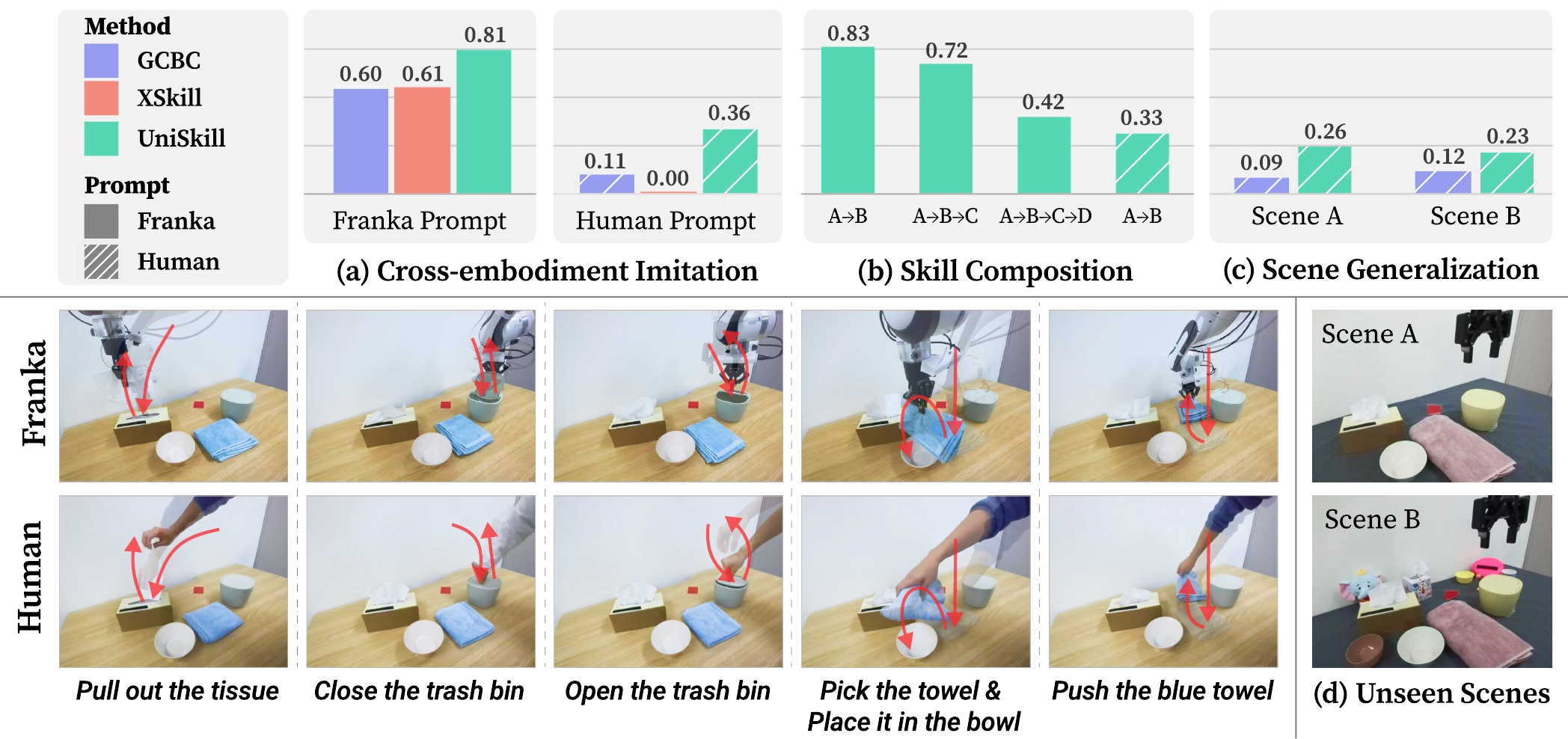}
    \caption{Overview of our tabletop experiments. (a) Average results on the tabletop benchmark using Franka and human prompts. (b) Results on skill composition using Franka and human prompts. \textbf{A:} \textit{Open the trash bin}, \textbf{B:} \textit{Pull out the tissue}, \textbf{C:} \textit{Pick the blue towel and place it in the bowl}, \textbf{D:} \textit{Close the trash bin}. (c) Results from human prompts evaluated on unseen environments in (d).}
    \label{fig:tabletop}
    \vspace{-1em}
\end{figure}

\paragraph{Evaluation Protocol.}
We conduct real-world experiments using a Franka robot across five tabletop tasks and three kitchen tasks, as well as simulation experiments on the LIBERO benchmark covering eight tasks.
Each task includes $100$ demonstrations and we fine-tune skill-conditioned policies separately for each benchmark.
As shown in \Cref{fig:tabletop}, tabletop evaluation uses two prompt types: \textbf{Franka} (same embodiment, held out from training) and \textbf{Human} (performed by humans, unseen during training).
For the kitchen benchmark, we also use \textbf{Anubis} prompts, collected from a custom Aloha-like robot~\citep{aloha, fu2024mobile} with an unseen embodiment, in an unseen environment (see \Cref{fig:kitchen}).
Performance is measured by the average success rate over three prompts per task with $20$ rollouts each.
Additional details on all benchmarks, including LIBERO, and robot hardware setups are provided in \Cref{sec:appendix_B}.

\paragraph{Baselines.}
We compare UniSkill with a goal-conditioned behavioral cloning policy (GCBC), which conditions on a goal image.
This baseline adopts the diffusion-policy architecture like ours, and is trained on goal images via hindsight relabeling~\cite{andrychowicz2017hindsight}. 
For a fair comparison, we condition the policy at inference on a sub-goal image $20$ frames ahead, matching the $20$-frame skill interval used by our skill-conditioned policy.
Apart from replacing the conditioning factor from a skill representation to a goal image, all other aspects remain identical in both training and inference.

We also compare against XSkill~\cite{xskill}, which learns a shared skill representation to enable cross-embodiment imitation through a self-supervised learning approach.
% Similar to UniSkill, it is conditioned on a prompt video to perform the task.
Unlike UniSkill, it requires a scene-aligned dataset, where human demonstrations are performed in the same environment and for the same task as the robot.
To support this, we collect an additional $100$ human demonstrations per task to train XSkill. Note that without this additional scene-aligned human video data, XSkill fails on all tabletop tasks (i.e., $0$ success).
More details on the baselines are provided in \Cref{sec:appendix_C}.

\subsection{Cross-Embodiment Imitation}
\begin{wrapfigure}{r}{0.45\textwidth}
    \centering
    \vspace{-2em}
    \includegraphics[width=\linewidth]{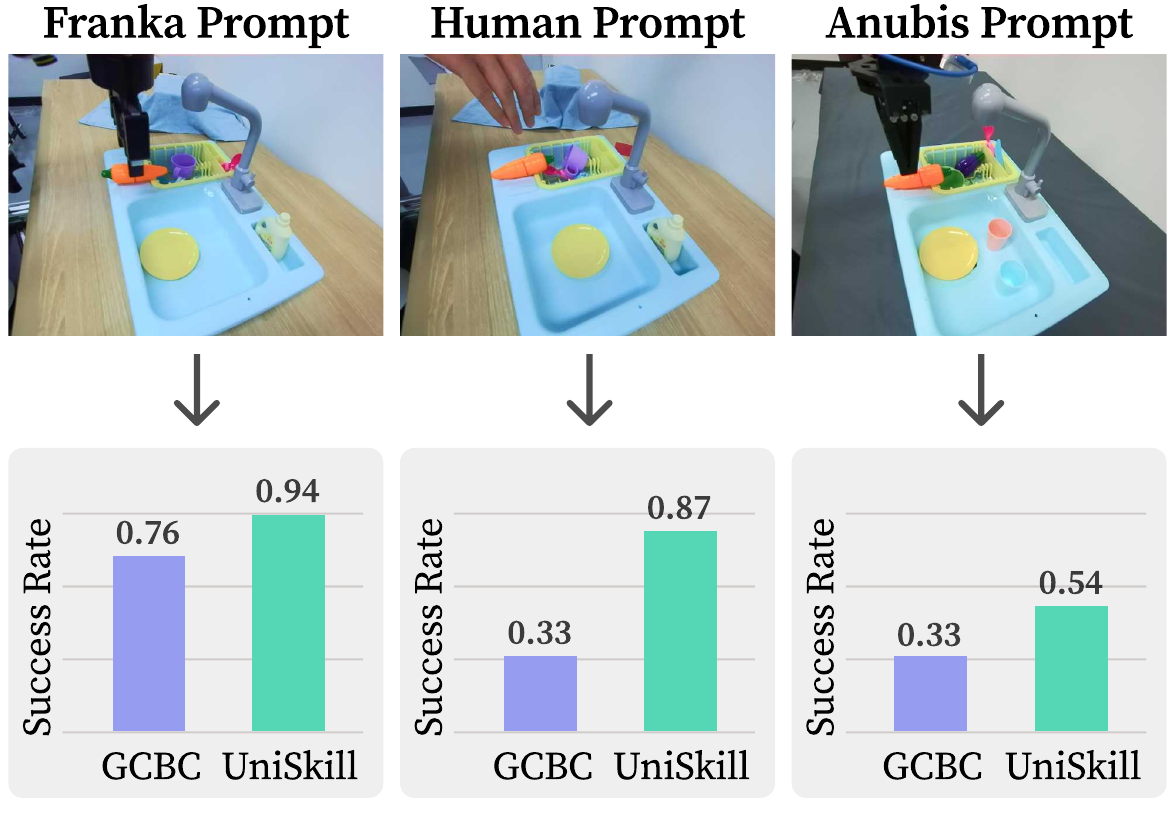}
    \caption{Results on the Kitchen benchmark
using Franka, Human, and Anubis (a different robot embodiment) prompts.}
    \label{fig:kitchen}
    \vspace{-1em}
\end{wrapfigure}
\Cref{fig:tabletop}(a) and \Cref{fig:kitchen} present the cross-embodiment imitation performance of UniSkill on real-world tabletop and kitchen benchmarks, and \Cref{fig:libero} shows the results on the LIBERO~\citep{libero} benchmark.
UniSkill consistently outperforms all baselines across both settings.
Notably, XSkill fails to imitate human videos,  even when trained directly on human demonstrations.
We attribute this to XSkill's clip-level contrastive learning objective, which does not effectively capture dynamics between frames.
On the other hand, UniSkill's image-editing based objective explicitly models temporal dynamics and generalizes well to both human and Anubis prompts, despite the former involving entirely different morphologies and the latter coming from an unseen robot in an unseen environment with novel objects.
This robustness highlights the embodiment-agnostic nature of UniSkill's skill representations enabled by large-scale video data including human videos.
Detailed task-wise results and additional results on the LIBERO benchmark are provided in \Cref{sec:appendix_A}.

\subsection{Cross-Embodiment Skill Representations}

\paragraph{Can UniSkill generalize to unseen, compositional tasks?}
During pre-training on large-scale video datasets, ISD compresses a motion pattern between two frames, allowing $\z_t$ to represent a low-level skill rather than a full task.
Although both UniSkill and GCBC are trained solely on demonstrations of individual tasks, we can assemble them at inference time to perform novel task combinations by leveraging the compositional nature of skills.
\Cref{fig:tabletop} (b) presents the results of task compositions in the tabletop benchmark.
While GCBC fails in all evaluations, UniSkill shows robust performance across all task combinations, even with human prompts.
This highlights that UniSkill captures a combinatorial space of diverse skills rather than overfitting to specific tasks, suggesting its potential scalability to a wide range of novel tasks.

\paragraph{Can UniSkill generalize to unseen environments?}
\label{sec:unseen}
\begin{wrapfigure}{r}{0.45\textwidth}
    \centering
    \vspace{-1em}
   \includegraphics[width=\linewidth]{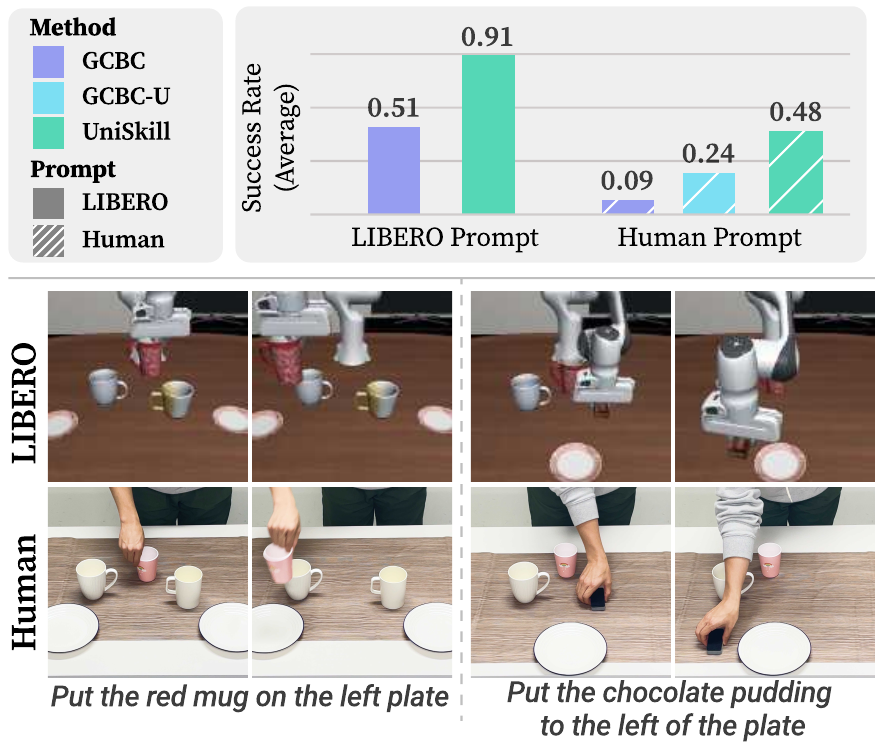}
    \caption{Average success rates for the LIBERO benchmark with \textbf{unseen human prompts (bottom)}. In human prompt videos, a human directly manipulates objects in a real-world environment similar to the LIBERO environment.}
    \label{fig:libero}
    \vspace{-2em}
\end{wrapfigure}
UniSkill leverages embodiment-agnostic skill representations to translate human video prompts into robot behaviors, despite not being trained on human prompts.
To further assess its generalization beyond embodiment, we evaluate Uniskill in two unseen environments: \textit{Scene A}, which alters the background and objects of the original tabletop benchmark, and \textit{Scene B}, which adds additional distractors into \textit{Scene A}, as illustrated in \Cref{fig:tabletop} (d).
\Cref{fig:tabletop} (c) shows that UniSkill achieves comparable performance across novel and visually modified scenes, which indicates that UniSkill is resilient to background and distractor variations.

To further validate scene-level generalization, we also conduct experiments in simulation, as shown in \Cref{fig:libero}.
Even when the human prompts come from entirely different environments, UniSkill is able to successfully infer and execute the intended task.
Further detail are provided in \Cref{sec:appendix_B}.

\begin{wraptable}{r}{0.35\textwidth}
    \vspace{-2em}
    \centering
    \resizebox{\linewidth}{!}{%
        \begin{tabular}{ccc|c}
        \toprule
            Droid       & Robot      & Human     & Avg\\
            \midrule
            \checkmark  &            &            & $0.56$      \\
            \checkmark  & \checkmark &            & $0.76$      \\      
            \midrule
            \checkmark  & \checkmark & \checkmark & $0.91$  \\
        \bottomrule
        \end{tabular}}
    \caption{Ablation studies on the impact of different training datasets, conducted on the LIBERO benchmark using robot prompts. \textbf{Robot:} Bridge and LIBERO. \textbf{Human:} Something-SomethingV2 and H2O.}
    \label{tab:ablation_dataset}
    \vspace{-2em}
\end{wraptable}
\paragraph{Does UniSkill benefit from incorporating human videos?} 
Using human videos for skill representation learning enables UniSkill to acquire diverse and transferable skills by leveraging large-scale video data.
As shown in \Cref{tab:ablation_dataset}, adding additional robot datasets (BridgeV2 and LIBERO) improves performance by $20\%$, while further incorporating large-scale human videos (Something-SomethingV2 and H2O) boosts it by an additional $15\%$.
This demonstrates that UniSkill benefits not only from scaling the robot dataset but also from using diverse human videos, highlighting the effectiveness of its embodiment-agnostic skill representation learning.

\begin{figure}[t]
    \centering
    \includegraphics[width=\linewidth]{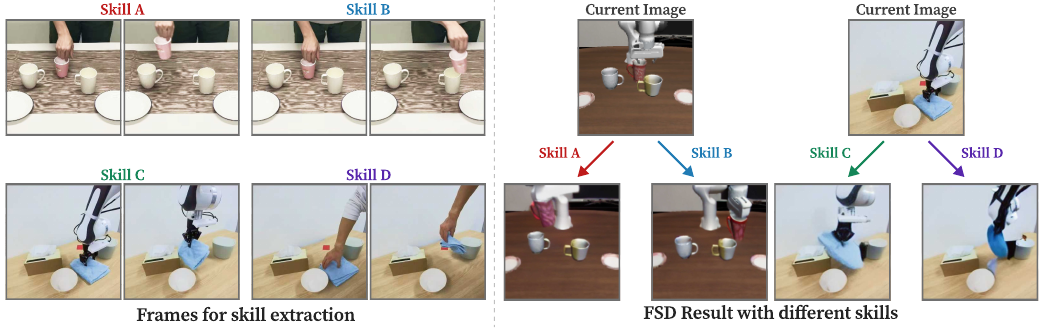}
    \caption{Qualitative results from FSD. The skill representation is extracted using ISD from each video prompt and conditioned on FSD to predict the future frame. (Left) Skills are extracted from two images using ISD. (Right) The predicted image generated by passing the current image and the extracted skill through FSD. Best viewed in color.}
    \label{fig:fsd}
    \vspace{-1em}
\end{figure}

\paragraph{Does UniSkill capture dynamic information?}
During skill representation learning, our image-editing based objective encourages the model to focus on dynamics changes between frames rather than static content, promoting the encoding of motion patterns into the skill representations.
To validate this, \Cref{fig:fsd} presents qualitative results of future frame prediction using FSD, conditioned on skill representations $\z_t$ from ISD.
Even when the current image is the same, the predicted future frame varies depending on the motion encoded in $\z_t$, despite the skills originating from different environments.
This confirms that the skill representation captures meaningful motion dynamics.
A comparison of dynamics awareness between UniSkill and prior works are provided in \Cref{sec:appendix_A2}.

\paragraph{Does UniSkill exhibit embodiment-agnostic properties?}
Unlike prior methods, UniSkill can learn cross-embodiment skill representations without requiring constrained human data, such as paired demonstrations with robots or matched environments.
\Cref{fig:fsd} shows that the predicted future frames from FSD preserve the original embodiment, even when the skill representation $\z_t$ is inferred from a different embodiment.
\begin{wrapfigure}{r}{0.4\textwidth}
    \vspace{-1em}
    \centering
    \includegraphics[width=\linewidth]{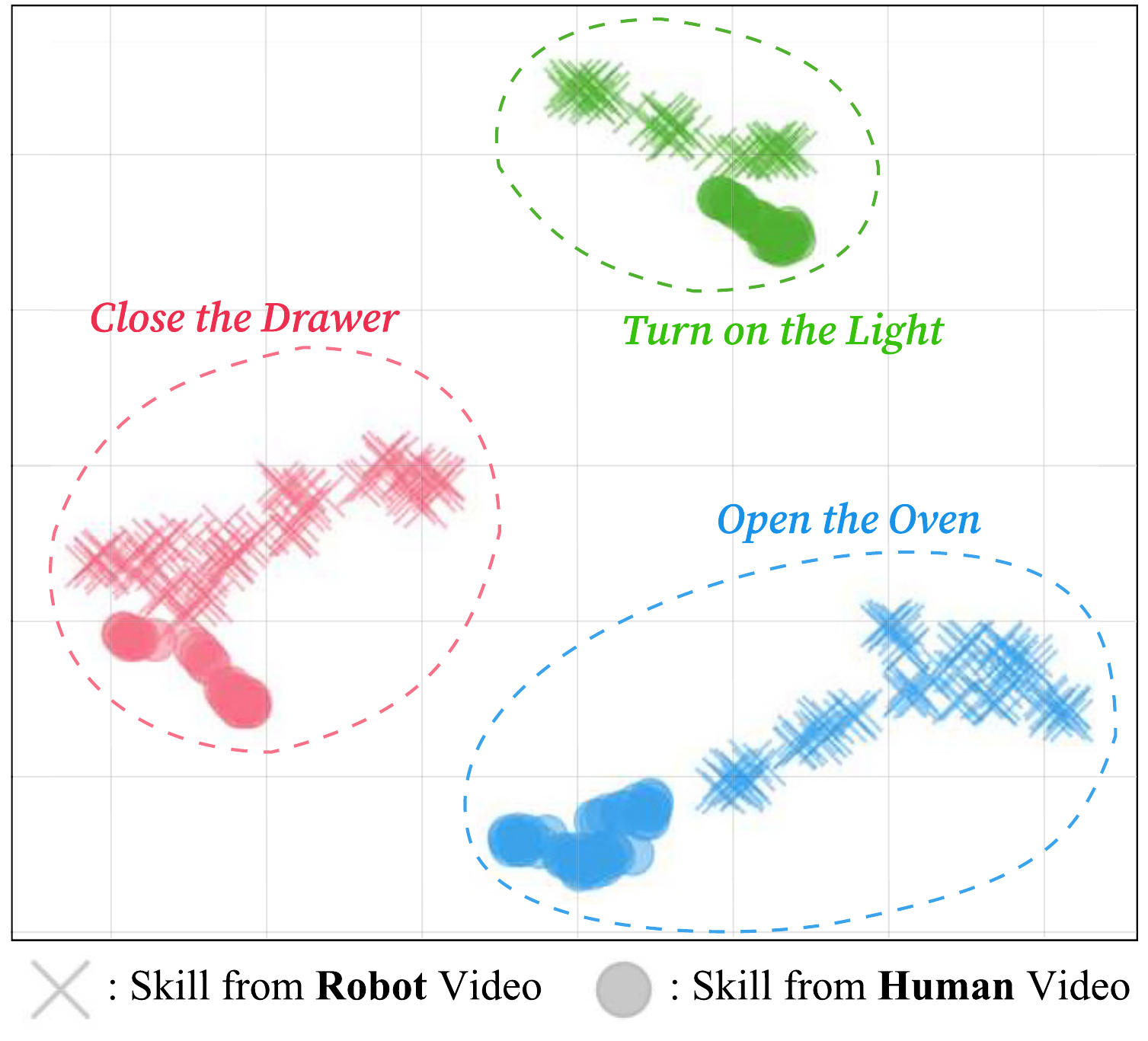}
    \caption{t-SNE visualization of UniSkill embeddings on the XSkill datasets. Circle markers indicate skill embeddings from human prompts, while cross markers represent those from robot prompts. Each color denotes a different skill. Example frames for each skill are shown in \Cref{sec:skillvis}.}
    \vspace{-2em}
    \label{fig:skill_vis}
\end{wrapfigure}
Notably, UniSkill preserves the correct embodiment when the current observation is from a simulation environment and the human prompt comes from a real-world setting.
Leveraging this property, we improve the original GCBC method, which suffers from performance degradation due to domain gaps between sub-goal images and the current observation.
As shown in \Cref{fig:libero}, we introduce \textbf{GCBC-U}, a variant that replaces GCBC's sub-goal image with an FSD-predicted frame (see details in \Cref{sec:appendix_A}), resulting in a $15\%$ performance improvement.

Additionally, \Cref{fig:skill_vis} presents a t-SNE visualization using the XSkill dataset~\citep{xskill}, which is not used for training.
The embeddings form task-specific clusters rather than embodiment-specific ones. The pattern indicates that the representation itself encodes embodiment-agnostic skills, since different tasks require different skill sets.
Together with the results from the real-world benchmarks, these findings highlight the embodiment-agnostic nature of UniSkill's skill representations.

\section{Conclusions}
\label{sec:conclusions}

In this paper, we propose UniSkill, a novel approach that successfully addresses cross-embodiment challenges without relying on a \pair{} during training.
Unlike prior works, UniSkill leverages unlabeled, large-scale video datasets spanning diverse embodiments to learn shared skill representations that generalize across embodiments.  
This enables impressive cross-embodiment imitation utilizing only the skill representations, without requiring additional inputs such as language instructions or goal images.  
UniSkill achieves comparable performance to existing methods and demonstrates the ability to mimic behaviors from video prompts, even when the prompts feature different embodiments.
Our results demonstrate that UniSkill effectively captures embodiment-agnostic dynamics information, allowing the policy to generalize across embodiments, making it a scalable solution for cross-embodiment imitation.

\section{Limitations}

UniSkill effectively encodes embodiment-agnostic dynamics into the skill representation, enabling policies to replicate behaviors from video prompts despite embodiment discrepancies. However, UniSkill has three primary limitations.

First, UniSkill relies on a fixed skill interval, which restricts its ability to adapt to varying execution speeds between human and robot demonstrations. Allowing for variable skill durations could improve its flexibility in handling differences in motion speeds across embodiments.  

Second, UniSkill struggles with videos that exhibit abrupt viewpoint changes, particularly in ego-centric human videos.  
Drastic visual shifts between consecutive frames hinder the extraction of coherent dynamic information, suggesting that improving robustness to such viewpoint variations is an important direction for future work.

Finally, UniSkill prefers prompt videos that exhibit robotic behaviors because its approach relies purely on motion imitation.
Without semantic cues such as task descriptions, having well-aligned behaviors in the prompts enhances performance.
A promising direction is to integrate UniSkill with vision-language-action (VLA) frameworks, where the VLA component provides high-level reasoning and UniSkill contributes precise motion representations for execution.

% The acknowledgments are automatically included only in the final and preprint versions of the paper.
\acknowledgments{This work was supported in part by the National Research Foundation of Korea (NRF) grant funded by the Korean government (MSIT) (RS-2024-00333634), and the Institute of Information \& communications Technology Planning \& Evaluation (IITP) grant funded by the Korea government (MSIT): the Artificial Intelligence Graduate School Program, Yonsei University (RS-2020-II201361) and the Global AI Frontier Lab International Collaborative Research (RS-2024-00469482 \& RS-2024-00509279).}

%===============================================================================

% no \bibliographystyle is required, since the corl style is automatically used.
\bibliography{references}  % .bib

\clearpage
\appendix

\section{Additional Experimental Results}
\label{sec:appendix_A}
\subsection{Detailed Results}
\subsubsection{Tabletop and Kitchen Benchmark}

\Cref{tab:main} presents the cross-embodiment capabilities of UniSkill's universal skill representation within the tabletop benchmark.
For Franka prompts, UniSkill achieves the highest performance on most tasks compared to the baselines.
While GCBC and XSkill show average success rate of $60\%$ and $61\%$, UniSkill maintains a minimum success rate of $75\%$, indicating consistently strong performance.
For human prompts, GCBC and XSkill fail to complete more than half of the tasks even once.
In contrast, UniSkill succeeds on most tasks and achieves an average success rate more than three times higher than the baselines.
This robustness, enabled by training on large-scale video data, demonstrates the generality of our skill representation across different embodiment, which is a central goal of our framework.

\Cref{tab:bridge} illustrates the performance on the kitchen benchmark.
UniSkill outperforms GCBC when evaluated with Franka prompts, which use an embodiment seen during both skill representation and policy learning.
Even with Anubis prompts, which involve an unseen robot embodiment, UniSkill still surpasses GCBC.
The performance gap is even more pronounced with human prompts, where UniSkill achieves more than twice the success rate of GCBC.
Notably, GCBC exhibits biased performance with unseen prompts, succeeding on only one out of three tasks for each type of different embodiment prompt.
This highlights GCBC's difficulty in handling demonstration videos from unseen embodiments.

\begin{table}[ht]
\begin{subtable}{0.58\linewidth}
\centering
    \resizebox{\linewidth}{!}{%
    \begin{tabular}{@{}clccc@{}}
    \toprule
        Prompt      & Task       & GCBC        & XSkill     & UniSkill     \\
        \midrule
        \multirow{6}{*}{Franka}
                    & \textit{Pull out the tissue}
                                 & $0.43$      & $0.42$     & $\mathbf{0.93}$             \\
                    & \textit{Push the blue towel}
                                 & $0.93$      & $\mathbf{0.97}$      & $0.75$             \\
                    & \textit{Close the trash bin}
                                 & $0.13$      & $0.58$     & $\mathbf{0.65}$             \\
                    & \textit{Open the trash bin}
                                 & $0.63$      & $0.80$     & $\mathbf{0.87}$             \\
                    & \textit{Pick the blue towel and place it in the bowl}
                                 & $0.62$      & $0.28$     & $\mathbf{0.85}$             \\
                    \cmidrule{2-5}
                    & Average
                                 & $0.60$      & $0.61$     & $\mathbf{0.81}$             \\
        \midrule
        \multirow{6}{*}{Human}
                    & \textit{Pull out the tissue}
                                 & $0.00$      & $0.00$     & $\mathbf{0.57}$             \\
                    & \textit{Push the blue towel}
                                 & $0.00$      & $0.00$     & $\mathbf{0.37}$             \\
                    & \textit{Close the trash bin}
                                 & $\mathbf{0.45}$      & $0.00$     & $0.25$             \\
                    & \textit{Open the trash bin}
                                 & $0.10$      & $0.00$     & $\mathbf{0.62}$             \\
                    & \textit{Pick the blue towel and place it in the bowl}
                                 & $0.00$      & $0.00$     & $0.00$             \\
                    \cmidrule{2-5}
                    & Average
                                 & $0.11$      & $0.00$     & $\mathbf{0.36}$             \\
    \bottomrule
    \end{tabular}
    }
    \caption{Tabletop}
    \label{tab:main}
\end{subtable}
\hfill
\begin{subtable}{0.38\linewidth}
    \centering
    \resizebox{\linewidth}{!}{%
        \begin{tabular}{@{}clcc@{}}
        \toprule
            Prompt      & Task       & GCBC        & UniSkill     \\
            \midrule
            \multirow{3}{*}{Franka}
                        & \textit{Put carrot on plate}
                                     & $0.58$& $\mathbf{0.90}$\\
                        & \textit{Turn faucet front to left}& $\mathbf{1.00}$& $\mathbf{1.00}$             \\
                        & \textit{Turn faucet front to right}& $0.70$& $\mathbf{0.93}$\\
                        \cmidrule{2-4}
                        &                                      Average& $0.76$& $\mathbf{0.94}$\\
            \midrule
            \multirow{3}{*}{Anubis}&                                      \textit{Put carrot on plate}& $0.00$& $0.00$\\
                        &                                      \textit{Turn faucet front to left}& $\mathbf{1.00}$& $0.83$\\
                        &                                      \textit{Turn faucet front to right}& $0.00$& $\mathbf{0.80}$\\
                        \cmidrule{2-4}
                        &                                      Average& $0.33$& $\mathbf{0.54}$\\
            \midrule
            \multirow{3}{*}{Human}
                        & \textit{Put carrot on plate}& $0.00$& $\mathbf{0.67}$\\
                        & \textit{Turn faucet front to left}& $\mathbf{1.00}$& $0.97$\\
                        & \textit{Turn faucet front to right}& $0.00$& $\mathbf{0.97}$\\
                        \cmidrule{2-4}
                        & Average
                                     & $0.33$& $\mathbf{0.87}$\\
        \bottomrule
        \end{tabular}
        }
        \caption{Bridge}
        \label{tab:bridge}
\end{subtable}
\caption{Real-world robot experiment results comparing UniSkill with baselines. Each task is evaluated using three prompts, and success rates averaged over 20 rollouts per prompt. (a) Results on the \custom{} benchmark using Franka and Human prompts. (b) Results on the kitchen benchmark using Franka, Anubis (a different robot embodiment), and Human prompts.}
\end{table}

\subsubsection{Task and Environment Generalization}

\begin{table}[h]
\begin{subtable}{0.43\linewidth}
    \centering
    \resizebox{\linewidth}{!}{%
    \begin{tabular}{clcc}
        \toprule
            Prompt      & Task             & GCBC                & UniSkill \\
            \midrule
           \multirow{3}{*}{Franka}
            & A + B
                        & $0.00$      & $\mathbf{0.83}$              \\
            & A + B + C
                        & $0.00$      & $\mathbf{0.72}$              \\
            & A + B + C + D
                        & $0.00$      & $\mathbf{0.42}$              \\
            \midrule
            Human & A + B & $0.00$      & $\mathbf{0.33}$              \\
        \bottomrule
    \end{tabular}
    }
    \caption{}
    \label{tab:skill}
\end{subtable}
\hfill
\begin{subtable}{0.55\linewidth}
    \centering
    \resizebox{\linewidth}{!}{
        \begin{tabular}{@{}llcccc@{}}
        \toprule
            \multirow{2}{*}{Task}   & \multicolumn{2}{c}{Scene A} & \multicolumn{2}{c}{Scene B}\\
                & GCBC        & UniSkill    & GCBC      & UniSkill  \\
            \midrule
            \textit{Pull out the tissue }
                & $0.00$      & $\mathbf{0.33}$      & $0.00$    & $\mathbf{0.23}$\\
            \textit{Push the blue towel}
                & $0.03$      & $\mathbf{0.18}$      & $0.00$    & $\mathbf{0.17}$\\
            \textit{Close the trash bin}
                & $0.05$      & $\mathbf{0.15}$      & $\mathbf{0.12}$    & $\mathbf{0.12}$\\
            \textit{Open the trash bin}
                & $0.02$      & $\mathbf{0.62}$      & $0.02$    & $\mathbf{0.57}$\\
            \textit{Pick the blue towel and place it in the bowl}
                & $\mathbf{0.35}$      & $0.00$      & $\mathbf{0.30}$    & $0.00$\\
            \midrule
            Average
                & $0.09$      & $\mathbf{0.26}$      & $0.12$    & $\mathbf{0.23}$\\
        \bottomrule
        \end{tabular}}
    \caption{}
    \label{tab:scene_robustness}
\end{subtable}
\caption{(a) Skill compositionality evaluation on the tabletop benchmark using Franka and human prompts. A composed task is considered successful only if all sub-tasks are completed. The sub-tasks are defined as follows: \textbf{A:} \textit{Open the trash bin}, \textbf{B:} \textit{Pull out the tissue}, \textbf{C:} \textit{Pick the blue towel and place it in the bowl}, \textbf{D:} \textit{Close the trash bin}. (b) Results on unseen scenes. Evaluation uses human prompts and follows the tabletop benchmark procedure.}
\end{table}
\Cref{tab:skill} presents detailed results for compositional tasks. For Franka prompts, performance decreases as task complexity increases, but UniSkill still achieves a $42\%$ success rate even when composing four tasks. In contrast, GCBC fails even on compositions of just two tasks. For human prompts, UniSkill achieves a $33\%$ success rate on two-task compositions, despite the prompts involving both an unseen embodiment and unseen tasks.
These results highlight the compositional nature of UniSkill's skill representation.
Although the composed tasks are not seen during policy learning, the skill-conditioned policy can still predict appropriate actions from the given skill representation.
This shows that even when tasks are novel, the policy can generalize across skills by executing actions aligned with the inferred motion patterns, resulting in successful behavior.

\Cref{tab:scene_robustness} reports per-task results for experiments in unseen environments. With human prompts, GCBC fails on most tasks, showing biased results with success only on one or two out of five tasks, and zero success on the rest. In contrast, UniSkill demonstrates generalization, successfully completing most tasks.
Similarly, in \Cref{fig:kitchen}, Anubis prompts are collected in unseen environments with a novel embodiment. While GCBC fails on all tasks under these conditions, UniSkill succeeds across them, as shown in \Cref{tab:bridge}. This is further supported by the results in \Cref{tab:libero}, where UniSkill achieves $48\%$ success with human prompts while GCBC reaches only $9\%$ (see \Cref{sec:sim}).
These results indicate that UniSkill is robust to scene variations in the prompt videos, consistently succeeding across the tabletop, kitchen, and simulation benchmarks.
They also support the conclusion that UniSkill can imitate behaviors from demonstration videos, regardless of the environment in which they were collected.

\begin{wrapfigure}{r}{0.4\textwidth}
    \centering
    \vspace{-1em}
   \includegraphics[width=\linewidth]{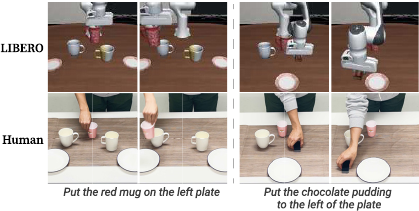}
    \caption{We created a prompt video in which a human directly manipulates objects after arranging them in a real-world environment similar to the LIBERO task. Here, we visualize only 2 out of the 8 tasks here for clarity.}
    \label{fig:libero_prompt}
\end{wrapfigure}

\subsection{Simulation Results on LIBERO}
\label{sec:sim}

\subsubsection{Evaluation Protocol}
We evaluate UniSkill on LIBERO~\cite{libero}, a benchmark designed for multi-task scenarios that features diverse object interactions, layouts, and tasks within a tabletop simulation environment using a Franka robot.
Our evaluation encompasses 8 tasks across 2 distinct scenes in LIBERO.
Detailed explanations of the tasks are provided in \Cref{sec:libero_task}.
Each task includes 50 expert demonstrations, which we use for policy learning.

For evaluation, one demonstration per task is selected as the Franka prompt.
To generate human prompts, we replicate the same tasks as those in the LIBERO benchmark.
However, due to the nature of the simulation environment, it is not possible to create human prompts that exactly match the simulation settings.
Instead, we align the number and positions of objects to closely resemble the LIBERO environment while ensuring a realistic human demonstrations, as shown in \Cref{fig:libero_prompt}.
As a result, the objects presented in the human prompts are largely novel and introduce previously unseen scenarios.
This implies that, while the behaviors demonstrated may align with those in the LIBERO benchmark, the semantic attributes of the objects and environment may differ.

To measure success rates, we use one demonstration per task and perform 20 rollouts per evaluation.

\subsubsection{Cross-Embodiment Skill}
\Cref{tab:libero} presents the evaluation results.
In the top section, UniSkill outperforms the baselines on robot prompts across all tasks.
Both GCBC and UniSkill are trained on expert demonstrations, but the result demonstrates the unique effectiveness of UniSkill's skill representations compared to raw pixel inputs.
This advantage is observed not only in real-world experiments but also in the simulation benchmark.

The bottom section of \Cref{tab:libero} evaluates cross-embodiment performance.
GCBC struggles to achieve meaningful performance when transitioning from LIBERO prompts to human prompts.
Especially, excluding tasks where both GCBC and UniSkill fail, GCBC's maximum success rate is $25\%$, whereas UniSkill's minimum success rate is $30\%$, surpassing GCBC's best result.

As shown in \Cref{fig:libero_prompt}, human prompts are not perfectly aligned with the simulation environment, making GCBC highly sensitive to such discrepancies.
Because GCBC predicts actions based on sub-goal images, large visual mismatches lead to extremly poor performance.
In contrast, UniSkill models the demonstrator's behavior from video prompts, allowing it to generalize across variations in object semantics.
This fundamental difference accounts for the significant performance gap between UniSkill and GCBC.

\begin{table}[t]
    \centering
    \small
        \begin{tabular}{ccccccccccc}
        \toprule
            \multirow{2}{*}{Prompt} & \multirow{2}{*}{Method}
                        & \multicolumn{4}{c}{\textbf{Scene1}}    & \multicolumn{4}{c}{\textbf{Scene2}}  &\multirow{2}{*}{Avg}                   \\
                        \cmidrule(l{2pt}r{2pt}){3-6} \cmidrule(l{2pt}r{2pt}){7-10}
            &           & Task1   & Task2   & Task3     & Task4     & Task5     & Task6     & Task7     & Task8  & \\
            \midrule
            \multirow{2}{*}{LIBERO} & GCBC
                        & $0.00$  & $0.55$  & $0.90$    & $0.30$    & $0.70$    & $0.95$    & $0.25$    & $0.35$   & $0.51$ \\
             & UniSkill
                        & $\mathbf{0.90}$  & $\mathbf{0.80}$  & $\mathbf{1.00}$    & $\mathbf{1.00}$    & $\mathbf{1.00}$    & $\mathbf{1.00}$    & $\mathbf{0.90}$    & $\mathbf{0.70}$  & $\mathbf{0.91}$  \\
            \midrule
            \multirow{3}{*}{Human}        & GCBC
                        & $0.05$  & $0.00$  & $0.25$    & $0.00$    & $0.25$    & $0.00$    & $0.05$    & $0.15$    & $0.09$\\
             & UniSkill
                        & $\mathbf{0.70}$   & $0.00$  & $\mathbf{0.80}$    & $0.00$    & $\mathbf{0.40}$    & $\mathbf{0.20}$    & $\mathbf{0.70}$    & $\mathbf{0.80}$ & $\mathbf{0.48}$\\
            \cmidrule{2-11}
             & GCBC-U
                        & $0.25$  & $\mathbf{0.10}$  & $0.50$    & $\mathbf{0.10}$    & $0.15$    & $0.15$    & $0.00$    & $0.65$    & $0.24$ \\
        \bottomrule
        \end{tabular}
    \vspace{1em}
    \caption{Performance comparison on the LIBERO simulation benchmark. For each task, one demonstration is used with 20 rollouts, and success rates are averaged to evaluate the performance.}
    \label{tab:libero}
\end{table}
\begin{figure}[t]
    \centering
    \includegraphics[width=\textwidth]{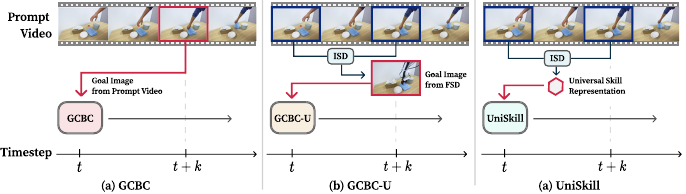}
    \caption{Comparison of the inference pipelines for GCBC, GCBC-U and UniSkill. All three methods use the same frame interval $k$. GCBC uses the $I_{t+k}$ frames as the sub-goal and predictions the actions required to achieve that state. In contrast, GCBC-U employs ISD and FSD to predict the sub-goal based on the current observation. UniSkill is directly conditioned on the skill representation from ISD rather than relying on a pixel-level goal condition.}
    \label{fig:baseline}
\end{figure}

\subsubsection{Improving GCBC with UniSkill}
Due to the embodiment-agnostic nature of its skill representation, UniSkill enables FSD to generate future frames that reflect the encoded motion while preserving the original embodiment.
This property can help resolve the embodiment mismatch issue in GCBC, where a sub-goal image from a human prompt not align with the robot's embodiment.

Building on this insight, we introduce \textbf{GCBC-U}, a variation of GCBC where the sub-goal image is replaced by one generated from FSD using UniSkill's skill representation.
As shown in \Cref{tab:libero}, GCBC-U significantly improves upon standard GCBC (from $9\%$ to $24\%$), despite the only change being the sub-goal input.
This highlights that the major limitation of GCBC lies in the embodiment discrepancy between the goal image and the target robot.
UniSkill's embodiment agnostic property effectively resolves this issue.

The overall inference pipelines of GCBC, GCBC-U, and UniSkill are illustrated in \Cref{fig:baseline}.

\subsection{Comparison with LAPA}
\label{sec:appendix_A2}
\begin{figure}[t]
    \centering
    \includegraphics[width=\textwidth]{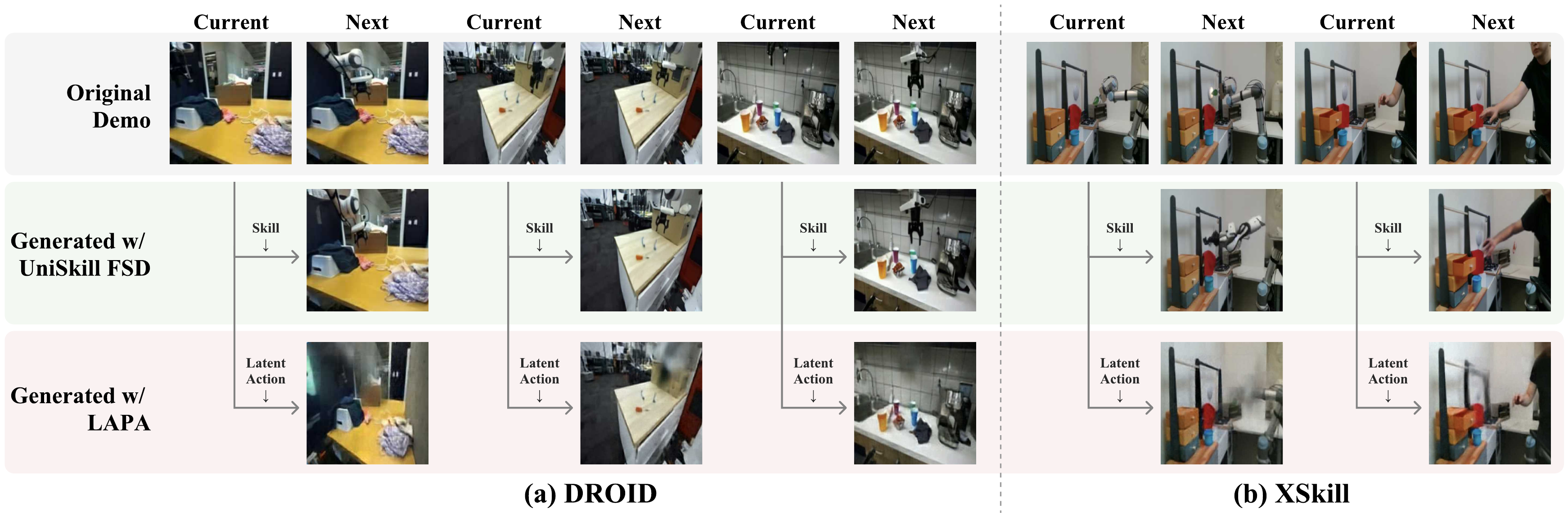}
    \caption{Comparison with Uniskill FSD and LAPA. A skill (UniSkill) or latent action (LAPA) was extracted between two frames, and the next frame was generated conditioned on the resulting vector. UniSkill FSD successfully reconstructs the video dynamics, while LAPA produces blurry images. (a): Result on DROID, (b): Result on XSkill.}
    \label{fig:fsd_comparison}
\end{figure}
Both LAPA~\cite{lapa} and UniSkill utilize diverse video datasets, including human demonstrations, to learn latent action or skill representations.
LAPA adopts Genie's~\cite{genie} transformer-based architecture trained with discrete latent actions.

In contrast, UniSkill employs an image editing~\cite{instructpix2pix} based pipeline to jointly train the Forward Skill Dynamics (FSD) and Inverse Skill Dynamics (ISD) models.
In this framework, the edited frame serves as the next frame, while the original frame is treated as the current frame.
This design encourages the ISD model to encode motion-specific features into the skill representation, rather than static features like background appearance.
As a result, UniSkill effectively captures the motion between two frames, which we refer to as a skill.

\Cref{fig:fsd_comparison} compares future frame predictions from LAPA and UniSKill, using latent embeddings produced by their respective inverse models.
We evaluate on two datasets: DROID~\cite{droid}, which is used for training, and XSkill~\cite{xskill}, which is unseen during training.

In \Cref{fig:fsd_comparison}(a), on the seen DROID dataset, LAPA generates blurry and less informative future frames, while UniSkill produces sharp and accurate predictions.
In \Cref{fig:fsd_comparison}(b), using the XSkill dataset--which is not used for training either method--only UniSkill accurately predicts the next frame, while LAPA continues to generate blurry outputs.
Notably, when tested on human demonstration videos, UniSkill predicts precise future frames based on the extracted skill representation, whereas LAPA merely reproduces the input frame, failing to model motion dynamics.
These results indicate that UniSkill's skill representation effectively captures dynamic changes between frames, while LAPA fails to do so.

\subsection{Additional Comparison with XSkill}
\subsubsection{Training XSkill on Large-Scale Datasets}
We observe low success rates for XSkill on our tabletop benchmark, even when using scene-aligned datasets--i.e., human and robot videos collected in the same environment and covering the same sub-tasks.
Although XSkill is typically used with scene-aligned data, it can be extended to unaligned, large-scale video datasets for skill discovery training, resembling the skill representation learning stage of UniSkill.
To enable a fairer comparison, we extend XSkill's training to include large-scale datasets and evaluate two variants that mirror UniSkill's training setup:

\begin{itemize}
    \item \textbf{XSkill-L} uses the same datasets as UniSkill for skill discovery. This includes robot datasets: Droid~\citep{droid}, Bridge~\citep{bridge}, and LIBERO~\citep{libero}  as well as human datasets: Something-Something V2~\citep{sthsth} and H2O~\citep{h2o}.
    \item \textbf{XSkill-A} builds on XSkill-L by additionally incorporating scene-aligned datasets collected in the tabletop environment.
\end{itemize}

\subsubsection{Progress-Based Success Metric}
In our primary evaluation, a task is marked as successful only if it is completed in full; otherwise, it is considered a failure.
While this binary metric is effective for comparing compact and robust methods, it cannot distinguish between completely failed attempts and those that achieve partial progress.

To address this, we introduce intermediate evaluation points for each task to define \textbf{partial success}.
Full task definitions are provided in \Cref{sec:real_benchmark}, and the partial success criteria are listed below:

\begin{itemize}
    \item \textbf{Pull out the tissue:} Move to the left - 0.3; reach above the tissue - 0.5; grasp the tissue - 0.7; fully pull out the tissue - 1.0.
    \item \textbf{Close the trash bin:} Move to the right - 0.3; reach above the trash bin - 0.5; reach behind the lid - 0.7; fully close the lid - 1.0.
    \item \textbf{Open the trash bin:} Move to the right - 0.3; reach above the trash bin - 0.5; fully open the lid - 1.0.
    \item \textbf{Pick the towel and place it in the bowl:} Move downward - 0.3; touch the towel - 0.5; lift the towel - 0.7; place the towel in the bowl - 1.0.
    \item \textbf{Push the blue towel:} Move downward - 0.3; touch the towel - 0.5; push without covering the mark - 0.7; push and cover the mark - 1.0.
\end{itemize}

\subsubsection{Effect of Dataset Scale and Alignment on XSkill}

\begin{wraptable}{r}{0.55\textwidth}
\centering
    \vspace{-1em}
    \resizebox{\linewidth}{!}{%
    \begin{tabular}{@{}clccc@{}}
    \toprule
        Prompt      & Task       & XSkill-L     &  XSkill-A     &UniSkill     \\
        \midrule
        \multirow{6}{*}{Franka}
                    & \textit{Pull out the tissue}
                                 & $0.17$&  $0.00$&$\mathbf{0.90}$\\
                    & \textit{Push the blue towel}
                                 & $0.03$&  $0.00$&$\mathbf{0.84}$\\
                    & \textit{Close the trash bin}
                                 & $0.61$&  $0.19$&$\mathbf{0.91}$\\
                    & \textit{Open the trash bin}
                                 & $0.57$&  $0.19$&$\mathbf{1.00}$\\
                    & \textit{Pick the blue towel and place it in the bowl}
                                 & $0.33$&  $0.06$&$\mathbf{0.90}$\\
                    \cmidrule{2-5}
                    & Average
                                 & $0.34$&  $0.09$&$\mathbf{0.91}$\\
        \midrule
        \multirow{6}{*}{Human}
                    & \textit{Pull out the tissue}
                                 & $0.63$&  $0.10$&$\mathbf{0.75}$\\
                    & \textit{Push the blue towel}
                                 & $0.20$&  $0.20$&$\mathbf{0.37}$\\
                    & \textit{Close the trash bin}
                                 & $0.50$&  $0.32$&$\mathbf{0.66}$\\
                    & \textit{Open the trash bin}
                                 & $0.50$&  $0.17$&$\mathbf{0.90}$\\
                    & \textit{Pick the blue towel and place it in the bowl}
                                 & $0.00$&  $0.04$&$\mathbf{0.21}$\\
                    \cmidrule{2-5}
                    & Average
                                 & $0.37$&  $0.17$&$\mathbf{0.58}$\\
    \bottomrule
    \end{tabular}}
    \caption{Real-world robot experiment results on tabletop benchmark using the progress-based metric. For both XSkill and UniSkill, each task is evaluated with three prompts, and success rates are averaged over five rollouts per prompt (UniSkill results are re-evaluated accordingly).}
    \label{tab:xskill_tabletop_progress}
    \vspace{-1em}
\end{wraptable}
\Cref{tab:xskill_tabletop_progress} presents comparison results on tabletop benchmark using the progress-based success metric.
The results show that UniSkill significantly outperforms both XSkill variants on Franka and human prompts.
UniSkill achieves near-perfect success ($91\%$) across diverse tasks with Franka prompts and also attains the highest success rate on human prompts.
In contrast, XSkill-L and XSkill-A perform poorly, despite being trained on the same or more datasets used for UniSkill's skill representation learning.
This highlights UniSkill's scalability with large-scale training, whereas XSkill struggles to scale effectively.
Notably, XSkill-L achieves only around $30\%$ success, corresponding roughly to the first stage of the progress metric.
This suggests that XSkill-L rarely completes tasks and often fails beyond the initial motion steps.

When comparing XSkill-L and XSkill-A, where the only difference is the inclusion of scene-aligned tabletop data, XSkill-L actually performs better, even though XSkill-A uses additional data.
This is likely because the added tabletop dataset contains only $1\mathrm{K}$ videos, which is much smaller than the large-scale training set of over $200\mathrm{K}$ videos.
As a result, the additional data has minimal effect on performance.
Moreover, this outcome reveals that XSkill's training becomes unstable when scaled to large and diverse datasets.

These results highlight a key limitation of XSkill's approach, which maps skill embeddings into a shared space using a fixed set of predefined prototypes.
While this mechanism allows mapping into a continuous skill representation, the limited number of prototypes restricts the model's ability to represent the wide variety of skills found in large-scale video datasets.
This limitation contributes to failures in completing full tasks and leads to instability during training.
For example, even under the progress-based metric, XSkill-L completely fails one task with a human prompt.

In contrast, UniSkill emphasizes motion by focusing on the dynamic parts of a video through an image-editing pipeline.
On the other hand, XSkill relies on learning objectives such as prototypes loss and time-contrastive learning, which are less effective at capturing motion patterns across video frames.
By directly encoding motion, UniSkill captures features that generalize well across diverse embodiments.
This allows it to demonstrate strong flexibility, even when responding to human demonstrations.

\begin{table}[h]
\begin{subtable}{0.35\linewidth}
    \centering
    \resizebox{\linewidth}{!}{%
        \begin{tabular}{cccc|c}
        \toprule
            Droid       & Robot      & XSkill      & Human     & Avg\\
            \midrule
            \checkmark  &            &            &            & $0.25$      \\
            \checkmark  & \checkmark &            &            & $0.19$      \\      
            \checkmark  & \checkmark & \checkmark &            & $0.19$  \\
            \checkmark  & \checkmark & \checkmark & \checkmark & $0.49$  \\
            \midrule
            \checkmark  & \checkmark &            & \checkmark & $0.48$  \\
        \bottomrule
        \end{tabular}}
    \caption{}
    \label{tab:ablation}
\end{subtable}
\hfill
\begin{subtable}{0.30\linewidth}
    \centering
    \resizebox{\linewidth}{!}{%
        \begin{tabular}{cc|c}
        \toprule
            Depth        & Augmentation &  Avg     \\
            \midrule
                         & \checkmark & $0.44$      \\      
             \checkmark  &            & $0.00$      \\  
             \checkmark  & \checkmark & $0.48$  \\      
            % \midrule
        \bottomrule
        \end{tabular}}
    \caption{}
    \label{tab:ablation_training}
\end{subtable}
\hfill
\begin{subtable}{0.31\linewidth}
\centering
    \centering
    \resizebox{\linewidth}{!}{%
    \begin{tabular}{@{}cccc@{}}
        \toprule
        \multicolumn{2}{c}{$k$} & \multicolumn{2}{c}{Prompt}\\
        Stage 1     & Stage 2        & LIBERO     & Human\\
        \midrule
        \multirow{2}{*}{$[1,20]$}   & $1$     & $0.19$        & $0.08$    \\
                                    & $20$    & $0.18$        & $0.05$    \\
        \midrule
        \multirow{2}{*}{$[20,40]$} & $20$    & $\textbf{0.91}$        & $\textbf{0.45}$    \\
                                    & $40$    & $0.79$        & $0.34$    \\
        \midrule
        \multirow{2}{*}{$[40,60]$} & $40$    & $0.80$        & $0.30$    \\
                                    & $60$    & $0.43$        & $0.19$    \\
        \bottomrule
    \end{tabular}}
    \vspace{-1mm}
    \caption{}
    \label{tab:skill_interval}
\end{subtable}
\caption{Ablation studies on the LIBERO benchmark using human video prompts. All experiments evaluate variations of UniSkill without relying on scene-aligned human-robot datasets. (a) Effect of training datasets The last row shows our method trained without the scene-aligned dataset (XSkill), yet achieving comparable performance. \textbf{Robot:} Bridge~\cite{bridge} and LIBERO~\cite{libero}. \textbf{Human:} Something-SomethingV2~\cite{sthsth} and H2O~\cite{h2o}. (b) Effect of training strategies. Both using augmentation and depth improve performance. (c) Effect of skill interval $k$. Stage 1 (skill representation learning) samples $k$ from a range, while Stage 2 (policy learning) uses a fixed interval.}
\end{table}

\subsection{Ablation Studies}
\label{sec:experiments:ablation}

All ablation studies are conducted in LIBERO benchmark with human prompt.
To conduct ablation studies, evaluation protocol is the same as simulation experiment.

\paragraph{Effect of Dataset for Pre-training.}
As shown  in \Cref{tab:ablation_dataset}, we already observe the importance of scaling up dataset size with robot prompts.
To further investigate the effect of pretraining datasets on cross-embodiment skill learning, we conduct ablation studies on various datasets with human prompt, with results presented in \Cref{tab:ablation}.
When large-scale human video datasets are used, performance more than doubles (from $19\%$ to $49\%$), highlighting the importance of including human data.
Interestingly, incorporating the XSkill dataset~\citep{xskill}, which is scene-aligned, does not lead to meaningful improvements--likely due to its relatively small size.
These findings suggest that the size and diversity of the dataset are more critical than whether it is scene-aligned.

\paragraph{Effect of Depth Prediction.}
\Cref{tab:ablation_training} demonstrates the importance of incorporating depth prediction.
Because our objective is to encode dynamics that transfer across embodiments, the representations should avoid overreliance on semantic cues that can entangle embodiment-specific features.
This issue is amplified in real-world setting, which are more heterogeneous than simulation environments.
As demonstrated in \Cref{fig:skillvis_full} (b), removing depth prediction results in a substantial drop in K-means clustering accuracy from $82.0$ to $31.7$, indicating reduced skill separation across embodiments.
A consistent trend appears on the Tabletop benchmark, where removing depth reduces K-means clustering accuracy from $62.8$ to $40.6$.
To mitigate this dependency, ISD explicitly incorporates depth predictions.
When depth prediction is not used, overall performance decreases, highlighting its importance in ensuring an embodiment-agnostic skill representation.

\paragraph{Effect of Augmentation.}
We evaluate the effectiveness of augmenting ISD inputs during policy learning.
As shown in \Cref{tab:ablation_training}, eliminating this augmentation leads to large drop in performance.
Because UniSkill encodes the dynamics of the video prompt, it is sensitive to changes in scene color and object appearance.
These results indicate that our augmentation strategy effectively mitigates these challenges. We use colorjittering for the augmentation.

\paragraph{Effect of Skill Interval.}
To validate our choice of the skill interval $k$, we conduct ablation studies on the LIBERO~\cite{libero} benchmark.
Training consists of two stages: skill representation learning and skill-conditioned policy learning.
During skill representation learning, we define a range for $k$ and sample a value from this range at each training iteration.
For policy learning, we use a fixed skill interval denoted by $k$.
As shown in \Cref{tab:skill_interval}, our default setting for $k$ achieves the best performance across both robot and human prompts.
\begin{wrapfigure}{r}{0.4\textwidth}
    \centering
    \vspace{-1em}
    \includegraphics[width=\linewidth]{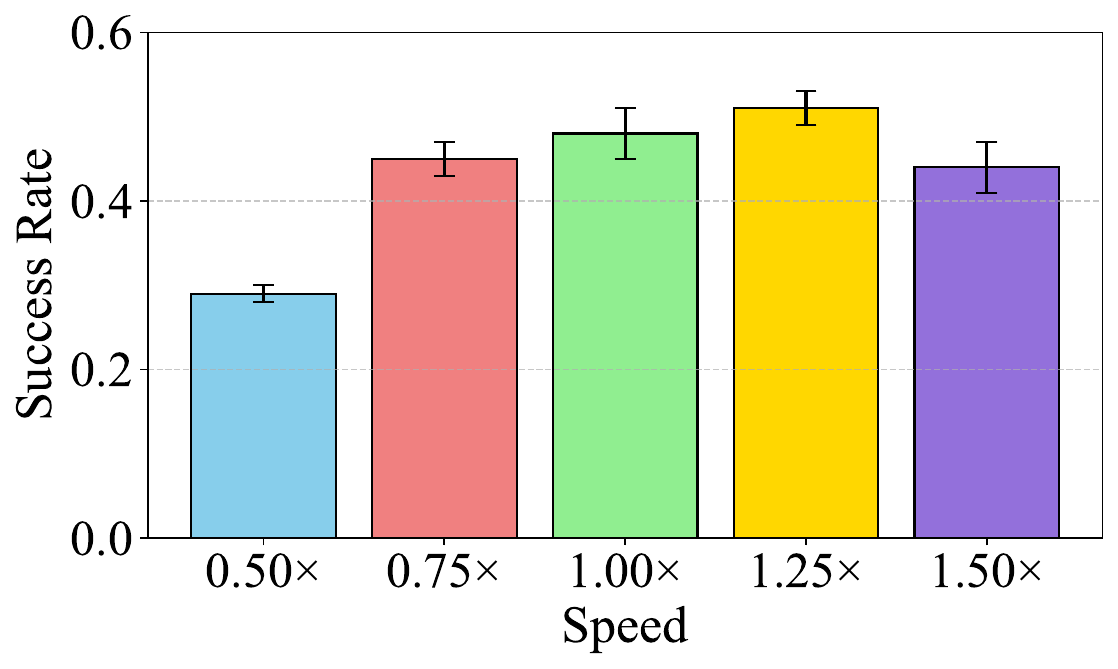}
    \caption{Ablation studies on camera speeds using human prompts on the LIBERO benchmark. Each success rate represents the average across all tasks in the benchmark.}
    \label{fig:speed}
    \vspace{-2em}
\end{wrapfigure}
When $k$ is too small, it becomes difficult to extract meaningful skill information between frames, leading to degraded performance.
On the other hand, if $k$ is too large, it may exceed the feasible execution horizon of a skill, which also harms performance, as seen when the policy learning interval is set to $60$.

\paragraph{Effect of Speed.}
In \Cref{fig:speed}, we evaluate the robustness of UniSkill by varying the video speed of the human prompts.
The best performance occurs at speeds of $1.00\times$ and $1.25\times$.
Notably, performance decreases as the speed slows down.
When the speed is too low, the encoded skill interval becomes short to capture meaningful action sequences.

\begin{table}[b]
\centering
    \begin{tabular}{@{}lcc@{}}
    \toprule
        \multirow{2}{*}{Task}   & \multicolumn{2}{c}{Scene B}\\
        & GCBC     & UniSkill$^\dagger$     \\
        \midrule
                     \textit{Pull out the tissue}
                                 & $0.42$     & $\mathbf{0.92}$             \\
                     \textit{Push the blue towel}
                                 & $\mathbf{0.43}$       & $0.33$             \\
                     \textit{Close the trash bin}
                                 & $0.08$     & $\mathbf{0.48}$             \\
                     \textit{Open the trash bin}
                                 & $0.52$     & $\mathbf{0.78}$             \\
                     \textit{Pick the blue towel and place it in the bowl}
                                 & $0.43$     & $\mathbf{0.90}$             \\
                    \midrule
                     Average
                                 & $0.38$     & $\mathbf{0.68}$             \\
    \bottomrule
    \end{tabular}
    \vspace{1em}
    \caption{Real-world robot experiments on the \custom{} benchmark comparing the performance of UniSkill and XSkill using robot prompts. $^\dagger$ indicates robot-only training, where only the DROID dataset is used for skill representation learning.}
    \label{tab:baseline_robotonly}
\end{table}
\subsection{Additional Analyses}
\subsubsection{Analysis of Skill-Conditioned Policy}
In addition to the embodiment-agnostic property of our skill representation, we further evaluate its effectiveness in policy learning.
To isolate this effect, we train the skill representation using only the DROID~\cite{droid} dataset.

\Cref{tab:baseline_robotonly} presents the results on the Scene B environment, which features a different background, different objects, and added distractors, as introduced in \Cref{sec:unseen}.
The strong performance in this unseen setting demonstrates the generalization capability and effectiveness of our skill representation, even when trained on a limited, robot-only dataset.

\begin{figure}[t]
    \centering
    \includegraphics[width=0.6\linewidth]{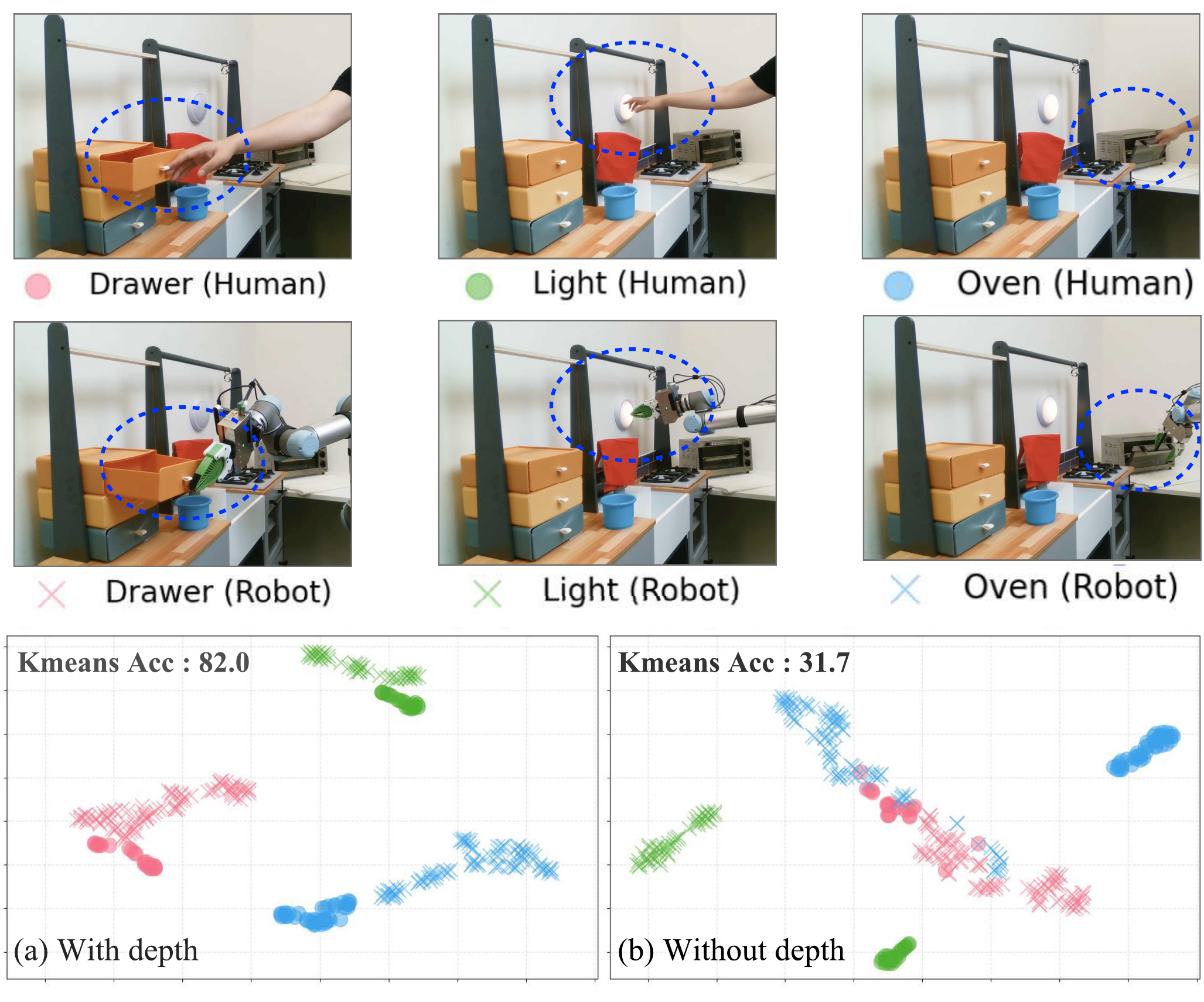}
    \caption{t-SNE visualization of UniSkill embeddings with and without depth on the XSkill dataset. Circle markers represent skill embeddings from human prompts, while cross markers represent those from robot prompts. Each color corresponds to a different task, with visual examples shown above for both human and robot executions.}
    \label{fig:skillvis_full}
\end{figure}

\subsubsection{Analysis of Cross-Embodiment Skill}
\label{sec:skillvis}
We further analyze UniSkill's ability to generalize across embodiments.
As discussed in \Cref{sec:experiments}, UniSkill's skill representations cluster by skill rather than embodiment.

To investigate this further, we visualize the t-SNE plots of skill embeddings with and without depth information.
As shown in \Cref{fig:skillvis_full}, the embeddings learned with depth are more compact and clearly separated by skill, while the embeddings without depth are more dispersed and overlapping.

\begin{wrapfigure}{r}{0.4\textwidth}
    \vspace{-1em}
    \centering
    \includegraphics[width=\linewidth]{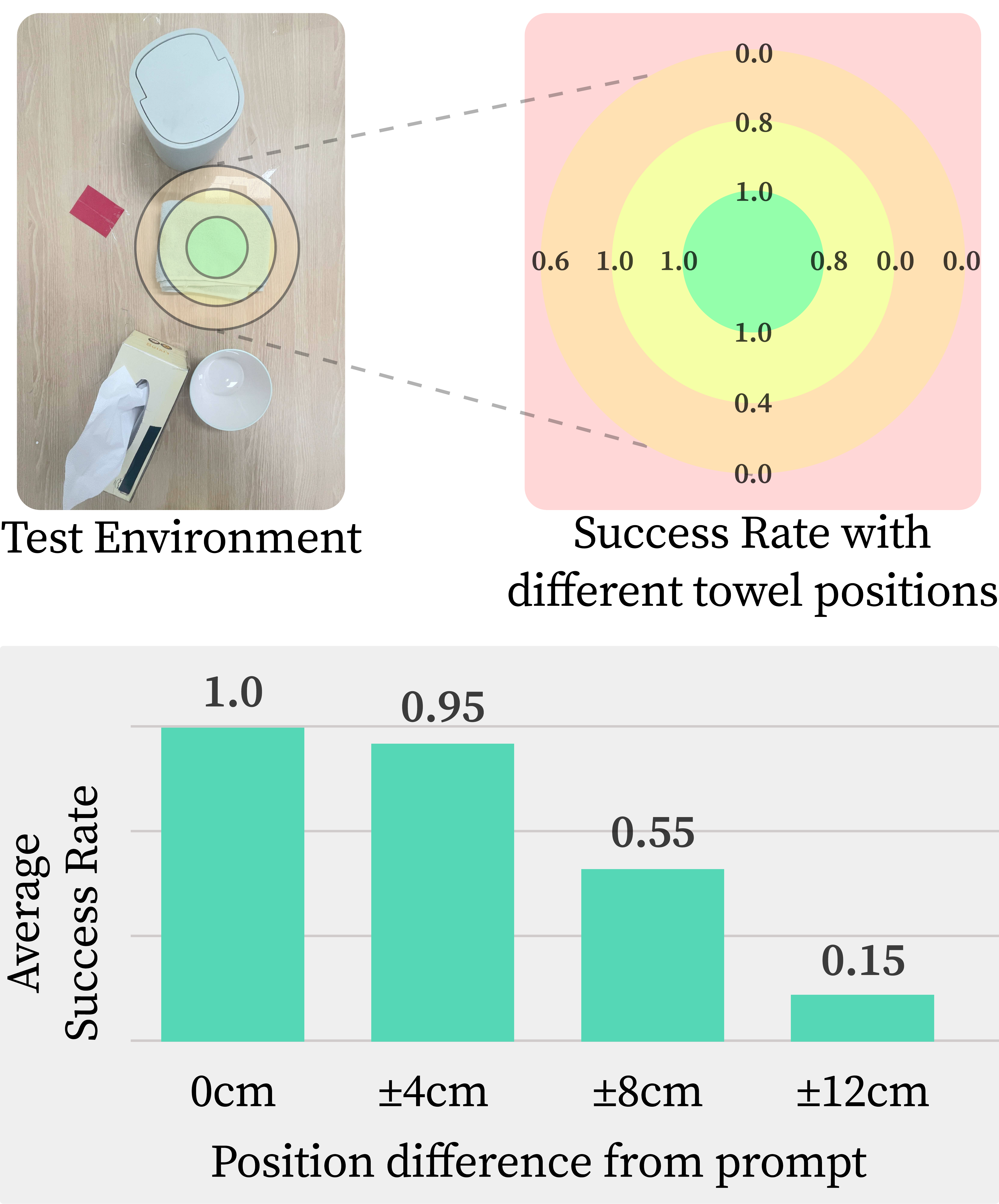}
    \caption{Visualization of the test environment and success rates across different initial towel positions. The towel's position is shifted up to $\pm12\mathrm{cm}$ from the prompt location to evaluate UniSkill's spatial sensitivity.}
    \label{fig:prompt_position}
    \vspace{-2em}
\end{wrapfigure}

Quantitatively, using K-means clustering with $K=3$, the depth-enabled model achieves higher clustering accuracy.
This suggests that incorporating depth improves the quality of the learned skill representation and enhances its embodiment-agnostic property.

\subsubsection{Analysis of Spatial Sensitivity of UniSkill}
UniSkill performs tasks by imitating motion patterns from demonstration videos.
As a result, the difference of positions of interacted objects between prompt video and test environment can influence task performance.
To investigate this, we design an experiment varying the position of the target object.

We select the task \textit{Push the blue towel} and modify the initial position of the towel in the evaluation environment relative to its position in the prompt video.
While the towel's position is already randomized during evaluation, we extend this variation to more extreme displacements to test the limits of spatial generalization.
As shown in \Cref{fig:prompt_position}, the towel's center is shifted by $0$ cm, $4$ cm, $8$ cm, and $12$ cm from the original prompt position.
The results show that as the position deviates further from the original, the success rate declines.

This drop in performance suggests that UniSkill, which emphasizes motion patterns over semantic cues, can be sensitive to spatial changes.
Nevertheless, it sill demonstrates a reasonable level of robustness, successfully completing the task across a range of varied object positions.

\section{Experiment Details}
\label{sec:appendix_B}
\subsection{Hardware Setup}
\begin{wrapfigure}{r}{0.4\textwidth}
    \vspace{0em}
    \centering
    \includegraphics[width=\linewidth]{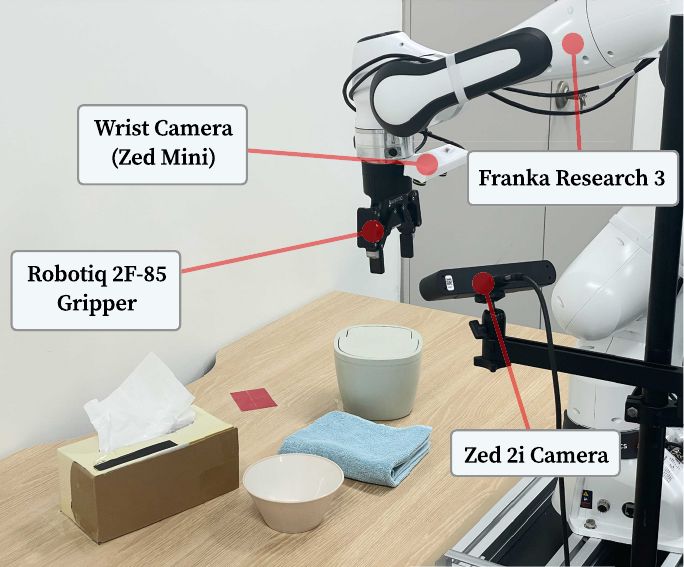}
    \caption{Our experiments are conducted in the DROID~\cite{droid} environment.}
    \label{fig:droid_setting}
    \vspace{-1em}
\end{wrapfigure}

We adopt the hardware configuration utilized in DROID~\cite{droid}.
Specifically, our setup comprises a Franka Research 3 robot arm paired with a 2F-85 Robotiq gripper.
For the camera setting, we use two cameras: a side camera and a wrist-mounted camera.
The side camera employed is the Zed 2i, while the wrist-mounted camera is Zed Mini.
Both cameras capture RGB images at a resolution of $720\times 1280$ at 15 Hz.
The overall settings are depicted in \Cref{fig:droid_setting}.

\subsection{Implementation Detail}
For pre-training, we initialize the FSD using the InstructPix2Pix model~\cite{instructpix2pix} and train the ISD from scratch.
For the visual encoder, we adopt the ResNet-18~\cite{resnet}, and for depth prediction, we utilize the pre-trained DepthAnythingV2 model~\cite{yang2024depth} without further training.
During pre-training, skill interval $k$ is randomly selected between $1.0s$ and $2.0s$, with the specific values determined by the frame rate of the video datasets.
The image resolution is set to $256\times 256$.
For policy learning, we employ diffusion policy~\cite{diffusionpolicy} as the policy network.
In real-world experiments, both UniSkill and GCBC are pre-trained on the DROID dataset~\cite{droid} and fine-tuned on the collected dataset.
During both training and inference, the skill interval $k$ is fixed at $20$ frames, the image resolution is $128\times 128$, and the action dimension is set to $7$.
The hyperparameters are provided in \Cref{sec:appendix_C}.

\subsection{Real-world Environments}
\label{sec:real_benchmark}

\subsubsection{Tabletop Benchmark}
For the \custom{} benchmark, we utilize four objects: a tissue box, bowl, towel, and trash bin.
The positions of the tissue box and trash bin are fixed, while the pose of the tissue varies.
For the bowl and towel, we define fixed regions and randomize their locations within those areas.
The towel's pose is also varied for each trial.
Additionally, we standardize the initial trajectory for each task to prevent the policy from becoming conditioned on specific initial movements.

\begin{figure}[ht] % Use figure* for 2-column figure
    \centering
    \includegraphics[width=\textwidth]{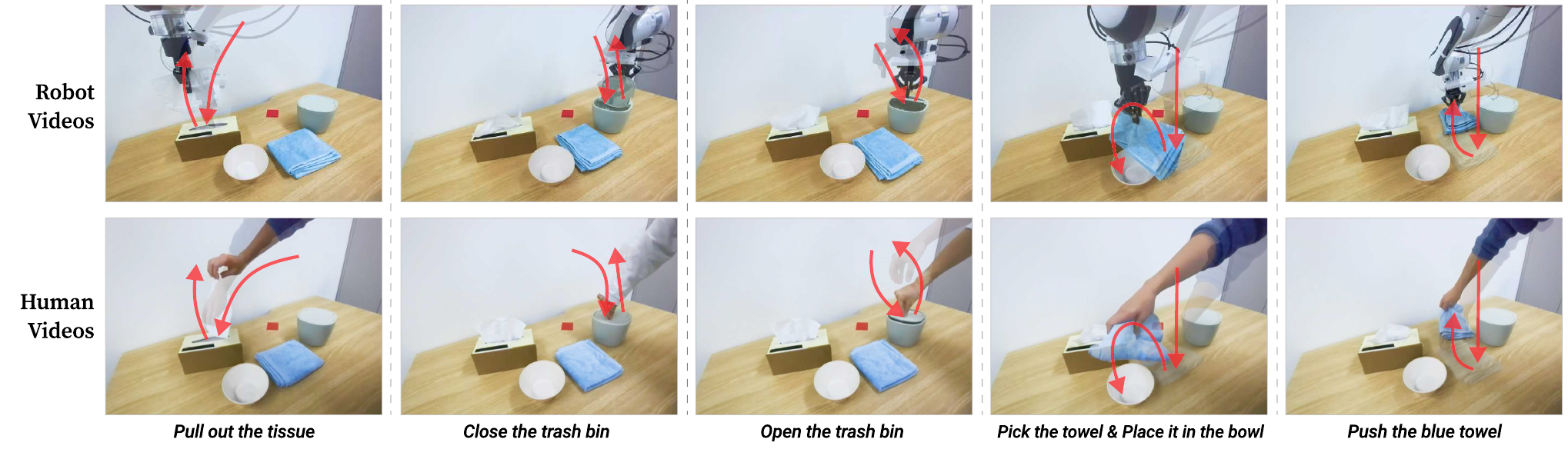}
    \caption{We designed five tasks within a real-world scene. The tasks were designed to share similar trajectories or involve the same objects across different tasks. This task setup allows for the evaluation of various actions, including push, pull, and pick-and-place.}
    \label{fig:droid_task}
\end{figure}

Task definitions are as follows:
\begin{itemize}
    \item \textbf{Pull out the tissue.} Pull out the tissue from the tissue box. \textit{Success Criterion}: The entire tissue is removed from the tissue box.
    \item  \textbf{Close the trash bin.} Close the lid of the trash bin. \textit{Success Criterion}: The lid is fully closed.
    \item  \textbf{Open the trash bin.} Open the lid of the trash bin. \textit{Success Criterion}: The lid is clearly opened without any partial closure.
    \item  \textbf{Pick the towel and place it in the bowl.} Pick up the blue towel and place it into the bowl. \textit{Success Criterion}: More than half of the bowl's area is covered by the towel.
    \item  \textbf{Push the blue towel.} Push the blue towel to the red mark. \textit{Success Criterion}: The red mark is entirely covered by the blue towel.
\end{itemize}

\Cref{fig:droid_task} shows the examples of task execution from both robot and human videos.
Note that human videos are not used during training.

\begin{figure}[ht]
    \centering
    \includegraphics[width=0.6\linewidth]{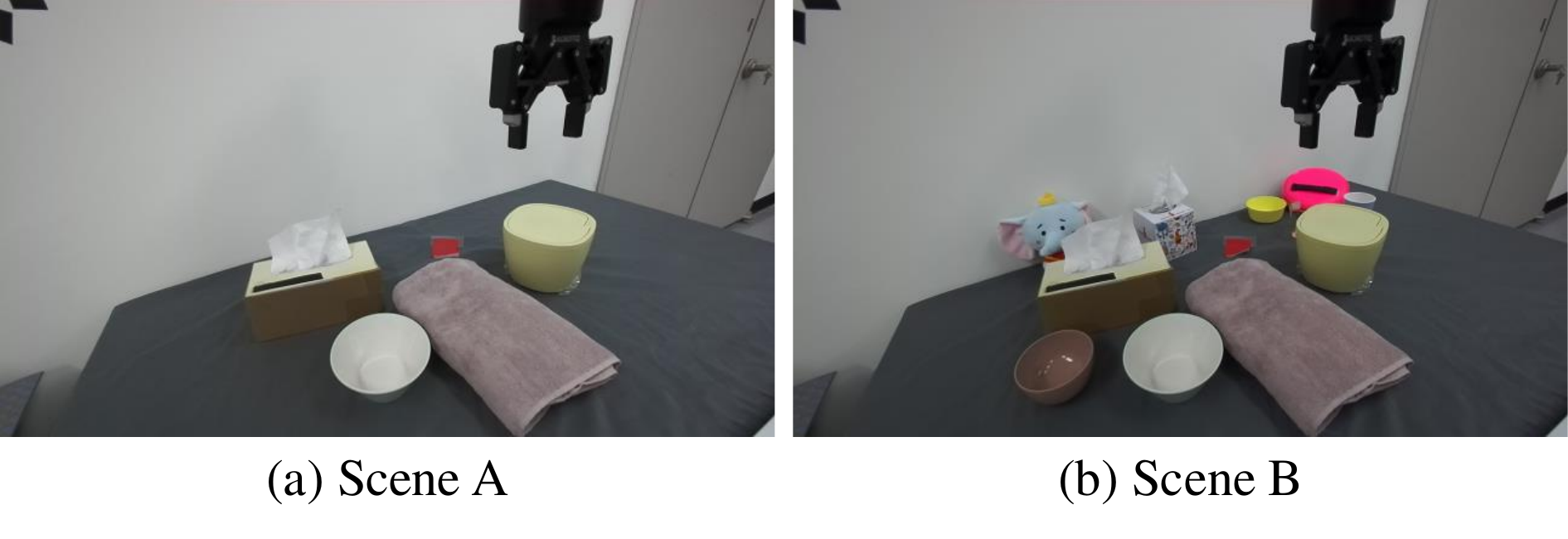}
    \caption{Unseen environments of \custom{} benchmark used for skill robustness evaluation.}
    \label{fig:unseen}
\end{figure}

For the skill robustness experiments in \Cref{sec:unseen}, we construct two new scenes.
\begin{itemize}
    \item Scene A: As shown in \Cref{fig:unseen}(a), we change the background (table), use a completely different towel in terms of shape, size, and color, and replace the trash bin and bowl with different colors.
    \item Scene B: As shown in \Cref{fig:unseen}(b), we introduce various distractor objects, including puppets, extra bowls, towels, and unrelated items, to increase visual complexity.
\end{itemize}

\subsubsection{Kitchen Benchmark}
For the kitchen benchmark, we employ a toy sink similar to that presented in the BridgeV2 dataset~\cite{bridge} and utilize three objects: faucet, carrot and plate.
The data collection process mirrors that of the \custom{} benchmark to maintain consistency across benchmarks.

\begin{figure}[ht]
    \centering
    \includegraphics[width=0.9\linewidth]{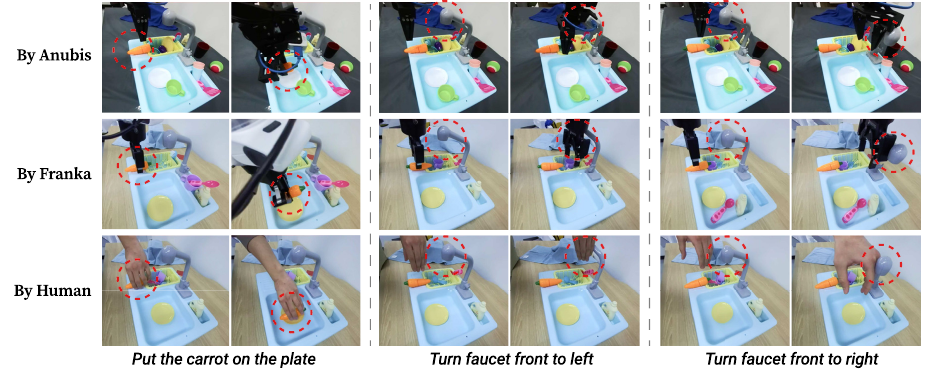}
    \caption{Prompt videos with different embodiments. To evaluate cross-embodiment imitation using UniSkill, we record prompt videos with 3 different embodiments. Prompt videos are recorded using a Anubis Robot (Top row), Franka arm (middle row) and a human hand (bottom row).}
    \label{fig:bridge_prompt}
\end{figure}

The task definitions are as follows:
\begin{itemize}
    \item \textbf{Turn faucet front to right.} Turn the head of faucet to the right direction. \textit{Success Criterion}: The faucet is moved to the right relative to its original position.
    \item \textbf{Turn faucet front to left.} Turn the head of faucet to the left direction. \textit{Success Criterion}: The faucet is moved to the left relative to its original position.
    \item \textbf{Put the carrot on the plate.} Pick up carrot and put it on the plate. \textit{Success Criterion}: The entire carrot is placed on the plate without any part touching the sink surface.
\end{itemize}

\Cref{fig:bridge_prompt} demonstrates examples of task executions using Franka, Anubis, and human embodiments.
Anubis is an unseen robot embodiment not included in skill representation learning, and its demonstrations are collected in an unseen environment with a completely different background.
\paragraph{Anubis Prompt.} 
Anubis is a custom-built robot inspired by Mobile ALOHA system~\citep{fu2024mobile}.
It is a mobile, bimanual robot with two 6-DoF arms, each equipped with a wrist-cam-mounted parallel gripper, and a 3-wheel omni chassis.  
As a non-commercial, custom-designed platform, Anubis does not appear in any existing robot datasets.
Therefore, it serves as a fully unseen embodiment in our evaluation.

\subsection{Simulation Environments}
\label{sec:libero_task}
For the LIBERO benchmark, we conduct experiments on four tasks within each of two distinct scenes, resulting in a total of eight tasks.
These tasks are predefined within the LIBERO simulation environment, and their success is automatically determined by the simulation system.

The task definitions are as follows:
\begin{itemize}
    \item \textbf{Task1.} put the red mug on the left plate.
    \item \textbf{Task2.} put the red mug on the right plate.
    \item \textbf{Task3.} put the white mug on the left plate.
    \item \textbf{Task4.} put the yellow and white mug on the right plate.
    \item \textbf{Task5.} put the chocolate pudding to the left of the plate.
    \item \textbf{Task6.} put the chocolate pudding to the right of the plate.
    \item \textbf{Task7.} put the red mug on the plate.
    \item \textbf{Task8.} put the white mug on the plate.
\end{itemize}

\section{Implementation Details}
\label{sec:appendix_C}

\subsection{Inference Procedure}
UniSkill’s policy is trained to produce actions conditioned on skill representations. At inference, given a video prompt, it derives a sequence of skill representations from the video using a sliding-window procedure.
Specifically, let the frames be $I_1,\dots,I_p$ and fixed interval $k$. For each $t=1,\dots,p-1$, we form the pair $(I_t,\; I_{\min(t+k,\,p)})$ and extract a skill $z_t$, so we obtain $z_1,\dots,z_{p-1}$. For the last k steps we use $I_p$ as the future frame.
For each timestep $t$, $z_t$ and observation $obs_t$ is then fed into the policy, which outputs the action executed in the environment.

\subsection{Skill Dynamics Modeling}
For skill dynamic modeling, UniSkill jointly trains the Inver Skill Dynamics (ISD) model and the Forward Skill Dynamics (FSD) model.
The hyperparameters used for training are listed in \Cref{tab:param_fsdisd}.

\begin{table}[ht]
    \begin{subtable}{0.4\linewidth}
        \centering
            \begin{tabular}{@{}ll@{}}
            \toprule
                
                Hyperparameter             & Value  \\
                \midrule
                Batch Size                 & $1024$    \\
                Training Epoch             & $50$      \\
                Learning Rate              & $1e-4$    \\
                $k$                        & $[20, 40]$\\
                skill dim                  & $256$     \\ 
                Optimizer                  & AdamW     \\
                Betas                      & $(0.9, 0.999)$ \\ 
                Weight Decay               & 0.01         \\
                Image Resolution           & $(256, 256)$    \\

            \bottomrule
            \end{tabular}
        \caption{}
        \label{tab:param_fsdisd}
    \end{subtable}
    \hfill
    \begin{subtable}{0.55\linewidth}
        \centering
        \resizebox{\linewidth}{!}{%
        \begin{tabular}{@{}ll@{}}
            \toprule
            Hyperparameter             & Value  \\
            \midrule
            \multirow{2}{*}{Batch Size} 
                                       & $128$ (DROID, tabletop, Bridge)   \\
                                       & $256$ (LIBERO)   \\         
            \midrule
            \multirow{4}{*}{Training Steps} 
                                       & 50000 (DROID) \\
                                       & 25000 (\custom{}) \\
                                       & 5000 (Bridge) \\       
                                       & 200000 (LIBERO) \\        
            \midrule
            Learning Rate              & $1e-4$    \\
            $k$                        & $20$\\
            Optimizer                  & Adam     \\
            Betas                      & $(0.9, 0.999)$ \\ 
            Weight Decay               & 0.01         \\
            Image Resolution           & $(128, 128)$    \\
            Crop Size                  & $(116, 116)$    \\
            Diffusion Model            & DDIM       \\
            Denoising Step             & $20$       \\
            Observation Horizon        & $2$            \\
            Prediction Horizon         & $16$           \\
            Action Horizon             & $8$            \\
            \bottomrule
        \end{tabular}}
        \caption{}
        \label{tab:param_policy}
    \end{subtable}
    \caption{Hyperparameters used in UniSkill: (a) FSD/ISD during pre-training and (b) policy learning.}
\end{table}

\subsubsection{Inverse Skill Dynamics Model}
For the Inverse Skill Dynamics model, we employ ResNet-18~\cite{resnet} as the visual encoder and utilize the pre-trained DepthAnythingV2-small model~\cite{yang2024depth} as the monocular depth estimator.
During pre-training, the depth estimator remains fixed, and only the visual encoder is trained to encode visual features.

To effectively capture spatial and temporal dependencies between frames $I_t$ and $I_{t+k}$, we integrate ST-Transformer blocks~\cite{st-transformer}.
Each ST-Transformer block comprises a spatial attention layer, a causal temporal attention layer, and a MLP layer.
The ISD model incorporates a total of eight ST-Transformer blocks, enabling robust encoding of dynamic interactions between the sampled frame pairs.
Prior to processing with the ST-Transformer blocks, the predicted depth maps are projected into depth features, which are then concatenated channel-wise with the visual features by the visual encoder for each timestep $t$ and $t+k$.

\subsubsection{Forward Skill Dynamics Model}
The Forward Skill Dynamics model adopts the architecture of InstructPix2Pix~\cite{instructpix2pix}, with a key modification: FSD is conditioned on the universal skill representation $\z_t$ instead of language instructions.
Consequently, while InstructPix2Pix freezes the text encoder and does not propagate gradients to it, we replace the text encoder with ISD and condition FSD on $\z_t$.
Additionally, the ISD receives gradient updates from FSD during training.
This adjustment ensures that FSD generates future frames based on the encoded motion patterns in $\z_t$, facilitating effective cross-embodiment imitation.

\subsection{Universal Skill-Conditioned Policy}
We employed a diffusion policy\cite{diffusionpolicy} as our policy architecture and utilized a codebase based on Robomimic \cite{mandlekarmatters} and DROID \cite{droid}. In the training process, we first resize all image observations to $128\times128$ and use a resnet \cite{resnet} visual encoder to extract visual features. These visual features are then concatenated with other observations to form a single vector. This observation vector is passed through an MLP to obtain a global condition. Typically, this global condition is fed into the Unet diffusion head to generate an action trajectory. However, in the case of our universal skill-conditioned policy, the global condition vector is concatenated with a universal skill representation before being processed by the diffusion Unet. 
We use an observation horizon of 2 and generated an action trajectory spanning 16 timesteps. During inference, the action prediction length is set to $8$, meaning that $8$ steps of actions are executed in an open-loop manner. For the diffusion model, we employ DDIM \cite{songdenoising} with $20$ denoising steps for action prediction.
The hyperparameters used for policy learning are reported in \Cref{tab:param_policy}.
\subsubsection{Real-World}
In real-world experiments, we use images from the ZED 2i camera and ZED Mini as image observations, along with the 3D Cartesian position of the gripper and the gripper state as proprioception.  ImageNet pretrained ResNet-50 \cite{resnet}  is used to encode the image observations.  During training, skills are extracted from the left ZED 2i camera images of expert trajectories at intervals of 20 timesteps. These skill representations are concatenated with the global condition for action denoising. During inference, skills are first extracted from the prompt video at intervals of 20 timesteps using ISD. These pre-extracted skills are then utilized as conditions corresponding to the current timestep for action denoising. Specifically, the skill relevant to the current timestep of prompt video is concatenated with the global condition, and the diffusion process is performed to predict the action trajectory.
\subsubsection{Simulation}
In the LIBERO \cite{libero} simulation setup, we use the robot's agent view and wrist view as image observations, with ResNet-18 \cite{resnet} as the visual encoder. For low-dimensional observations, we include the end-effector's orientation, position, gripper states, and joint states. Similar to the real-world experiment, we extract skills only from the agent view and not from the wrist view during training. These extracted skills are then concatenated with the global condition to predict actions. During inference, we extract skills from the prompt video with skill interval $20$ and use the skill corresponding to the current timestep as a condition for action prediction.

\subsection{Goal-conditioned Behavioral Cloning}
\label{sec:baseline_gcbc}

The policy architecture of a goal-conditioned behavioral cloning policy (GCBC) employs the same diffusion policy as the Universal skill-conditioned policy, with all components being identical except for the conditioning. In the Universal skill-conditioned policy, the global condition is concatenated with the Universal skill representation for denoising, whereas in GCBC, the global condition is concatenated with the goal image feature for denoising. The goal image feature is obtained by passing the corresponding view image through the same visual encoder used for encoding observations.

During training, the goal image is sampled from the expert dataset using hindsight relabeling and is concatenated with the global condition for action prediction. During inference, to maintain consistency with the Universal skill-conditioned policy setup, the image from 20 timesteps ahead in the prompt video is used as the sub-goal image for the current timestep.

\subsubsection{Real-World}
The real-world setup for GCBC is largely consistent with that of the Universal Skill-Conditioned Policy. Similar to the UniSkill setup, the image from wrist camera is not used as the goal image, and the image from the ZED 2i camera is utilized. During inference, the image from $20$ timesteps ahead in the prompt video is used as the goal image for action denoising. 
\subsubsection{Simulation}
The LIBERO \cite{libero} simulation setup is also identical to that of the Universal Skill-Conditioned Policy. The LIBERO simulation agent-view is used as the goal image, hence the goal image feature is obtained by passing the goal image through the visual encoder for agent-view observations.

\subsection{XSkill}
\label{sec:baseline_xskill}
XSkill~\citep{xskill} proposes a cross-embodiment skill representation using a feature clustering approach.
While it does not require strictly paired datasets with identical motions between humans and robot demonstrations, it sill imposes certain constraints.
Specifically, XSkill requires human demonstration videos for training, and those videos must be recorded in the same environment and cover the same tasks as the target robot setup.
\begin{wraptable}{r}{0.45\textwidth}
\vspace{-0em}
\begin{subtable}{\linewidth}
    \centering
        \begin{tabular}{@{}ll@{}}
        \toprule
            
            Hyperparameter             & Value  \\
            \midrule
            Video Clip Length $l$      & $8$    \\
            Sample Frames $T$          & $100$      \\
            Sinkhorn Iterations        & $3$    \\
            Sinkhorn Epsilon           & $0.03$    \\
            Prototype Loss Coef        & $0.5$     \\ 
            Prototype Loss Temperature & $0.1$     \\ 
            TCN Loss Coef              & $1$        \\
            TCN Positive Window $w_p$  & $16$        \\
            TCN Negative Window $w_p$  & $16$        \\
            TCN Positive Samples       & $1$        \\
            TCN Temperature $\tau_{\mathrm{tcn}}$              & $0.1$        \\
            Batch Size                 & $20$         \\
            Training Iteration         & $500$         \\
            Learning Rate              & $1e-4$    \\
            Optimizer                  & $\mathrm{ADAM}$    \\
        \bottomrule
        \end{tabular}
    \caption{}
    \label{tab:param_xskill_skill}
\end{subtable}
\vfill
\begin{subtable}{\linewidth}
    \centering
        \begin{tabular}{@{}ll@{}}
        \toprule
            Hyperparameter             & Value  \\
            \midrule
            Observation Horizon        & $2$        \\
            Observation Dimension      & $7$        \\
            Action Dimension           & $7$        \\
            Batch Size                 & $128$         \\
            Training Iteration         & $150$         \\
            Learning Rate              & $1e-4$    \\
            Optimizer                  & $\mathrm{ADAM}$    \\
        \bottomrule
        \end{tabular}
    \caption{}
    \label{tab:param_xskill_skill_composing}
\end{subtable}
\caption{Hyperparameters used for XSkill: (a) Skill Discovery and (b) Skill Transfer Composing.}
\vspace{-3em}
\end{wraptable}
To meet these requirements, we additionally collect 100 human demonstrations per task in the same scene as the target evaluation setup.

Since the official XSkill codebase\footnote{https://github.com/real-stanford/xskill} does not include complete inference code or training configurations for real-world, we re-implement the method for real-world experiments, following the paper and available codebase as closely as possible. Unless otherwise noted, we keep the hyperparameters from the original XSkill setup without modifications, including the number of skill prototypes.
For the reported results, we train XSkill three times and report the best performance among the runs.
The training configurations used for XSkill are reported in \Cref{tab:param_xskill_skill} and \Cref{tab:param_xskill_skill_composing}.

\section{Failure Cases}
\begin{figure}[ht] 
    \centering
    \includegraphics[width=\textwidth]{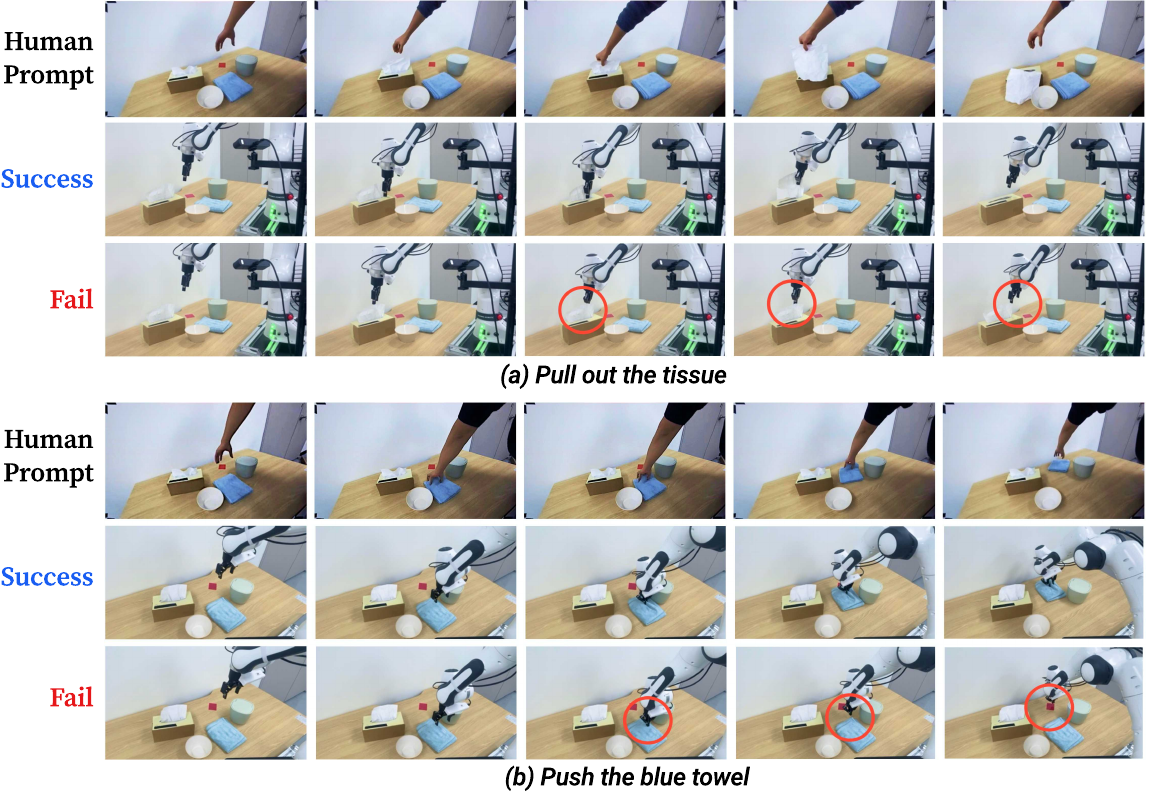}
    \caption{Analysis of failure cases for UniSkill on the tabletop tasks \textit{Pull out the tissue} and \textit{Push the blue towel}. In these cases, the primary failure mode is inaccurate contact with the target object.}
    \label{fig:failure}
\end{figure}

We analyze failure cases in the real-world tabletop benchmark.
\Cref{fig:failure} displays both successful and unsuccessful rollouts derived from the same human video prompts.
A common failure mode is the inability to make proper contact with the target objects.
Although our skill representation encodes motion patterns and the robot faithfully follows the demonstrated trajectory, it does not adapt when object interaction fails.

In failure case (a), for example, the gripper slightly retracts toward the tissue, attempts to grasp it, but only opens without securing the object.
In failure case (b), the gripper descends to grasp the towel but misses, resulting in a failure to secure the towel even though the robot proceeds to push toward the desired location.

These observations suggest that while UniSkill effectively replicates the demonstrated trajectory, further enhancements in object interaction may yield additional performance gains.

\end{document}